\RequirePackage{fix-cm}
\documentclass[twocolumn]{svjour3}          
\smartqed  

\usepackage{bbding}
\usepackage{graphicx}
\usepackage{amssymb}
\usepackage[cmex10]{amsmath}
\usepackage{array}
\usepackage{url}
\usepackage{color}
\usepackage[tight,footnotesize]{subfigure}
\usepackage{natbib}
\usepackage{hyphenat}
\usepackage[pdftex,
CJKbookmarks=true, bookmarksnumbered=true, bookmarksopen=true,
colorlinks=true,
pdfborder=001,
citecolor=blue, linkcolor=blue, anchorcolor=red, urlcolor=black
]{hyperref}

\usepackage{bbm,dsfont}
\usepackage{indentfirst} 
\usepackage{graphicx}
\usepackage{amsmath}
\usepackage{amssymb}
\usepackage{booktabs}
\usepackage{bm}

\usepackage[dvipsnames]{xcolor} 
\usepackage{caption}
\usepackage{subcaption}
\usepackage{multirow}
\usepackage{tabularx}
\usepackage{graphicx}
\usepackage{adjustbox}
\usepackage{amssymb}
\usepackage{pifont}
\usepackage{algorithmic}
\usepackage[ruled,vlined]{algorithm2e}
\usepackage[skip=2pt]{caption}

\usepackage{marvosym}
\usepackage[normalem]{ulem}

\newcommand{\tsb}{\textsubscript}

\newcommand{\vA}{{\mathbf{A}}}
\newcommand{\vB}{{\mathbf{B}}}
\newcommand{\vC}{{\mathbf{C}}}

\newcommand{\vG}{{\mathbf{G}}}
\newcommand{\vH}{{\mathbf{H}}}

\newcommand{\vJ}{{\mathbf{J}}}

\newcommand{\vQ}{{\mathbf{Q}}}

\newcommand{\vW}{{\mathbf{W}}}

\newcommand{\calT}{{\mathcal{T}}}

\newcommand{\bbR}{{\mathbb{R}}}

\newcommand{\ie}{\textit{i.e.}}

\newcommand{\tsp}{\textsuperscript}
\newcommand{\tdr}{\textsuperscript{\textdagger}}

\newcommand{\str}{\textsuperscript{*}}
\newcommand{\uar}{\(\uparrow\)}
\newcommand{\dar}{\(\downarrow\)}
\newcommand{\pms}[1]{\textsubscript{\(\pm\){#1}}}

\newcommand{\use}{\(\surd\)}

\newcommand{\tblu}[1]{\textcolor{blue}{#1}}

\newcommand{\calO}{{\mathcal{O}}}

\journalname{International Journal of Computer Vision}
\definecolor{darkmag}{rgb}{0.55,0,0.55}

\begin{document}
\begin{sloppypar}

\title{Mamba-CL: Optimizing Selective State Space Model in Null Space for Continual Learning}


\author{De Cheng \and
        Yue Lu \and
        Lingfeng He \and
        Shizhou Zhang \and
        Xi Yang \and
        Nannan Wang \and
        Yanning Zhang \and
        Xinbo Gao        
}


\institute{
De Cheng~*(Equal contribution) \at Xidian University, Xi'an 710071, China \\
\email{\href{dcheng@xidian.edu.cn}{dcheng@xidian.edu.cn}}
\and
Yue Lu~*(Equal contribution) \at Northwestern Polytechnical University, Xi'an 710129, China \\
\email{\href{zgxd@mail.nwpu.edu.cn}{zgxd@mail.nwpu.edu.cn}}
\and
Lingfeng He \at Xidian University, Xi'an 710071, China \\
\email{\href{lfhe@stu.xidian.edu.cn}{lfhe@stu.xidian.edu.cn}} 
\and
Shizhou Zhang~\Letter~(Corresponding author) \at Northwestern Polytechnical University, Xi'an 710129, China \\
\email{\href{szzhang@nwpu.edu.cn}{szzhang@nwpu.edu.cn}}
\and
Xi Yang \at
Xidian University, Xi'an 710071, China \\
\email{\href{yangx@xidian.edu.cn}{yangx@xidian.edu.cn}}
\and
Nannan Wang \at
Xidian University, Xi'an 710071, China \\
\email{\href{nnwang@xidian.edu.cn}{nnwang@xidian.edu.cn}}
\and
Yanning Zhang \at Northwestern Polytechnical University, Xi'an 710129, China \\
\email{\href{ynzhang@nwpu.edu.cn}{ynzhang@nwpu.edu.cn}}
\and
Xinbo Gao \at
Chongqing University of Posts and Telecommunications, Chongqing 400065, China \\
\email{\href{gaoxb@cqupt.edu.cn}{gaoxb@cqupt.edu.cn}}
}

\date{Received: date / Accepted: date}

\maketitle

\begin{abstract}
Continual Learning (CL) aims to equip AI models with the ability to learn a sequence of tasks over time, without forgetting previously learned knowledge. Recently, State Space Models (SSMs), particularly the Mamba model, have achieved notable success in computer vision. Building on the strengths of SSMs, this study explores leveraging the Mamba model for CL. Therefore, we introduce Mamba-CL, a framework that continuously fine-tunes the core SSMs of the large-scale Mamba foundation model by updating parameters orthogonal to the feature subspace of previous tasks. In selective SSMs, forgetting can be understood as changes in the state-update and output-generation processes activated by previous-task inputs. Accordingly, Mamba-CL uses the selective scan structure to identify the parameter updates that should be constrained, while retaining directions for learning new tasks. This approach derives conservative sufficient conditions that provide a tractable approximation to the general output-consistency requirement for each SSM module across previous and current tasks, thereby mitigating \emph{catastrophic forgetting}. Specifically, we achieve this goal by deducing the overall consistency constraints on four key time-invariant parameters in the Mamba model, streamlining its recurrent state-space structure and non-linear discretization process in SSM.
In practice, we apply the null-space projection to efficiently implement the orthogonality within Mamba model. Extensive experiments on four class-incremental benchmarks demonstrate the effectiveness of Mamba-CL for anti-forgetting, achieving superior performances to state-of-the-art methods.
Code is available at \url{https://github.com/zugexiaodui/mamba-cl}.

\end{abstract}

\section{Introduction}
Continual Learning (CL) aims to enable AI models to learn new tasks without forgetting previously acquired knowledge while processing sequentially arrived data, which is a capability essential for adapting to dynamic real-world environments.
However, deep models face significant challenges in mitigating \emph{catastrophic forgetting} when adapting to new data.
This makes it difficult to achieve a balance between learning new knowledge and retaining prior knowledge, known as the \emph{stability-plasticity dilemma}.

Existing CL methods can be broadly categorized into three types: \emph{regularization-based} methods~\cite{EwC,OWM,nullspace2021}, \emph{rehearsal-based} methods~\cite{aljundi2019gradient,hu2022curiosity,zhao2021memory}, and \emph{network expansion} methods~\cite{gao2022efficient,hung2019compacting,yoon2019scalable}.
Among regularization-based approaches, subspace projection methods have gained attention for their promising performance in both convolutional neural networks (CNNs)~\cite{nullspace2021} and vision transformers (ViTs)~\cite{lu2024visual}.
Specific methods such as OWM~\cite{OWM} and NSCL~\cite{nullspace2021} have been effectively applied to CNNs, and VPT-NSP\tsp{2}~\cite{lu2024visual} has demonstrated strong performance for ViTs.
Recently, State Space Models (SSMs)~\cite{2021s4model, 2022s5model}, particularly the Mamba model~\cite{2023mamba}, have achieved notable success in both sequence modeling and computer vision, with applications like De-focus Mamba~\cite{2024defocusmamba} and VisionMamba \cite{2024visionmamba}.
Building on the strengths of SSMs in vision tasks, this study explores the potential of the Mamba model for CL through subspace projection.

To this end, we propose Mamba-CL, a method that fine-tunes the SSMs along a direction orthogonal to the subspace spanned by those features from previous tasks during training Mamba foundation model in a new task.
By doing this, the interference with previously learned tasks can be minimized, thus effectively reducing \emph{catastrophic forgetting} in CL.
Although gradient orthogonal projections, which theoretically prevent forgetting, have been widely studied for CNNs, directly applying these methods to the Mamba structure introduces significant challenges due to: 1) the higher-order recurrent state-space structure, and 2) the non-linear discretization mechanism in SSMs.
These factors make gradient orthogonal projection in SSMs considerably more complex than in CNNs, posing challenges for continual training of the Mamba-based foundation model.

In the proposed Mamba-CL framework, we aim to continually fine-tune the core SSM blocks within the Mamba foundation model, by updating parameters in the direction orthogonal to the subspace spanned by input features from previous tasks.
To achieve this, we formulate an objective that ensures output consistency across tasks in SSMs.
Given the challenges posed by the higher-order recurrent state-space structure and the nonlinear discretization process, we decompose the analysis into three key components.
For each component, we establish sufficient conditions on parameter updates to satisfy the consistency objective.
Consequently, we demonstrate that this objective can be achieved by enforcing orthogonality constraints on four independent, time-invariant parameters in SSMs.
Specifically, consistency is maintained if: 1) updates of the discretization step parameter (\(\bm{\delta}\)) and the output projection matrix (\(\mathbf{C}\)) remain orthogonal to the subspace spanned by input features, 2) updates of the state transition matrix (\(\mathbf{A}\)) is orthogonal to the discretization step parameter, and 3) updates of the input projection matrix (\(\mathbf{B}\)) are orthogonal to the product of the discretization step parameter and input features.
These derived conditions provide a theoretically motivated and tractable approximation for reducing interference with previously learned knowledge through gradient orthogonal projection.

Orthogonal subspace constraints are well suited to selective SSMs because forgetting can be understood as changes in the state-update and output-generation processes activated by old-task inputs. Since these inputs typically span only a limited feature subspace, restricting new-task updates to orthogonal directions can reduce changes to old-task responses while retaining capacity for learning new tasks. Selective scan specifies the input-dependent processes to be protected, whereas the derived orthogonal constraints restrict the corresponding parameter updates from altering those processes.

In practice, we implement the derived sufficient \emph{consistency conditions} by a null-space-based approximation solution~\cite{nullspace2021,lu2024visual}.
It enables efficient gradient orthogonal projections for the continual adaptation of the Mamba foundation model.
Moreover, we introduce a balance factor to relax the orthogonality constraints so that we can make a good trade-off between stability and plasticity.
We validate our Mamba-CL approach, which fine-tunes the Mamba foundation model with null-space projections, across several class-incremental benchmarks, including 10-split and 20-split ImageNet-R, 10-split CIFAR-100, and 10-split DomainNet.
The proposed Mamba-CL demonstrates solid effectiveness in mitigating \emph{catastrophic forgetting}, even in challenging long-sequence CL scenarios.

The main contributions can be summarized as follows:
\begin{itemize}
\item We formulate continual adaptation of selective SSMs from an output-preservation perspective, linking selective scan with orthogonal subspace constraints that limit changes to the state-update and output-generation processes activated by previous-task inputs.
\item We theoretically derive four sufficient consistency conditions for the SSM block within the Mamba model. Based on these conditions, we introduce a null-space-based approximation to efficiently implement gradient orthogonal projections, enabling continual adaptation of the Mamba foundation model.
\item Extensive experimental results demonstrate the effectiveness of our approach in anti-forgetting on four class-incremental benchmarks with diverse experimental settings, and our approach achieves superior performances to state-of-the-art methods.
\end{itemize}

\section{Related Works}
\subsection{State Space Models (SSMs)}
State Space Models (SSMs) are proposed to model the long-range dependencies \cite{2021s4model, 2022s5model} in Natural Language Processing (NLP).
\cite{2021s4model} first introduces a Structured State-Space Sequence (S4) model based on the control theory, outperforming previous models in accuracy and speed.
Mamba~\cite{2023mamba} proposes a data-dependent SSM block and designs a generic language model for various NLP tasks.
Recently, the SSM structure has been investigated in computer vision tasks.
Several attempts \cite{2023mamba_for_movie, 2023mamba_for_vedio, 2024umamba} design SSM-based vision models for various vision tasks, such as video understanding and biomedical image segmentation.
VisionMamba \cite{2024visionmamba} first proposes a vision backbone consisting of bidirectional Mamba blocks, which achieves superior performance and memory efficiency over vision transformers.
VMambda \cite{2024vmamba} further proposes Visual State-Space blocks and a 2D Selective Scan module based on the SSM to efficiently gather context for images.
De-focus Mamba \cite{2024defocusmamba} devises bandpass filters and enhances training strategies to broaden the attention range, further improving the performance of the SSM-based model in vision tasks.
Despite the difference in the design of visual Mamba blocks among existing models, they all rely on the SSM for context modeling.
Inspired by the advancement of SSM models in vision tasks, this study seeks to harness the potential of SSM for CL.

\subsection{Typical Continual Learning Techniques}
\textbf{Rehearsal-based methods} \cite{iCaRL,rwalk2018,replay3} allow storing a small subset of training samples during the learning process and utilize these stored samples with new task data for model training.
The key challenge for such methods lies in determining which samples to retain to effectively preserve knowledge from previous tasks.
For example, \cite{iCaRL} employ a herding strategy that aims to make the stored samples’ feature mean as close as possible to the mean of all training samples’ features.
\cite{rwalk2018} propose selecting samples that are either near the classification decision boundary or have high output probability values, as they are deemed more critical for preserving decision information.
Additionally, \cite{replay3} introduce a gradient-based sample selection method that maximizes the diversity of samples in the replay buffer through a greedy algorithm.
Although rehearsal-based methods have achieved state-of-the-art continual learning performance, they introduce additional storage overhead and pose potential privacy concerns.

\textbf{Network expansion-based methods} \cite{expansion3,gao2022efficient,yoon2019scalable} retain the parameter branches of previous tasks and expand the network by adding new branches for each new task.
For instance, the DER method proposed by \cite{expansion3} freeze previously learned feature extractors when facing new tasks and expands the backbone network by introducing new feature extractors.
To explore more optimal strategies for adding new branches, \cite{gao2022efficient} propose a dynamic network expansion method based on neural architecture search, which achieves neuron-level control to minimize the expansion cost.
However, as the number of tasks increases, the continuous expansion of branches leads to increased inference overhead, ultimately raising the deployment cost in practical applications.

\textbf{Regularization-based methods} \cite{EwC,LossPlasticity2024Dohare,OWM,nullspace2021} mitigate catastrophic forgetting by constraining parameter updates to prevent significant changes in important parameters.
These techniques are mainly categorized into two categories.
Parameter importance estimation approaches estimate the importance of parameters for previous tasks and applies constraints to their updates.
For example, \cite{EwC} incorporate parameter constraints into the loss function, identifying optimal parameters that balance learning new tasks while maintaining performance on previous tasks.
\cite{regularization2} estimate parameter importance by computing the sensitivity of the loss function to changes in each parameter during training.
\cite{LossPlasticity2024Dohare} improve model plasticity for new tasks by selectively reinitializing less important parameters when learning new tasks.

Orthogonal subspace projection approaches leverage orthogonal projection \cite{OWM, nullspace2021, 2021gpm, 2023dualgpm, deng2021flattening} to constrain gradient update directions when learning new tasks.
Specifically, parameters are updated in the subspace orthogonal to the previous feature space.
Through the orthogonality, the features from old tasks can remain unchanged after learning new tasks, thereby theoretically enhancing the stability of models.
For instance, \cite{OWM} project the gradients of new tasks into a subspace orthogonal to the input feature space, ensuring that network outputs for previous task samples remain unchanged while preserving useful directions for learning new tasks.
Subsequent studies \cite{nullspace2021, 2021gpm, 2023dualgpm} introduce improved implementations of orthogonal projection to enhance computational efficiency and achieve better plasticity-stability trade-off.
Orthogonal projection-based methods can theoretically eliminate feature drift, thus effectively mitigating catastrophic forgetting.
However, existing research on this approach has been limited to linear layers such as convolutional and fully-connected layers.

\subsection{Pre-trained Model-based Continual Learning}
Recently, pre-trained model-based CL methods utilizing pre-trained ViTs have achieved remarkable performance~\cite{l2pCVPR22,dualpromptECCV22,codaCVPR23,cpromptCVPR24, RevisitingClassIncremental2025Zhou, he2026harnessing, he2025ckaa}, where a small number of learnable prompts are introduced for the adaptation to downstream tasks.
These methods mainly focus on innovations in prompt pool \cite{l2pCVPR22} architectures.
The L2P framework proposed by \cite{l2pCVPR22} first introduces a cross-task shared key-prompt storage mechanism that embeds task-specific knowledge through a two-stage optimization process.
To address the limitations in prompt selection accuracy inherent in this framework, subsequent research has advanced in three key directions.
1) Prompt selection mechanism: To improve prompt selection precision, \cite{dualpromptECCV22} enhance L2P by introducing expert prompts for each task. \cite{codaCVPR23} further incorporate an attention-weighted mechanism to enable end-to-end training.
2) Dynamic prompt generation: \cite{2024ConvPrompt} propose a convolution-based method to dynamically generate task-specific prompts, while \cite{evopromptAAAI24} introduce evolutionary parameter memory to facilitate cross-task prompt evolution.
\cite{WhenPromptbased2023Tang} propose a learnable generator that transforms the prompt pool architecture from a static storage system to a dynamic generation system.
3) Training strategy optimization: \cite{cpromptCVPR24} employ a random prompt selection strategy during training to enhance inference robustness.
\cite{AttriCLIPNonIncremental2023Wang} employ language modality information, such as class text descriptions, to train text prompts that guide the continual updating of image prompts.

In addition to improving the prompt pool, some methods have incorporated regularization techniques into continual learning based on pre-trained ViTs.
\cite{pgpICLR24} first propose to update the visual prompts in the direction orthogonal to the previous input feature subspace.
\cite{hetask} further project the learnable update into two distinct low-rank subspaces for better trade-off between knowledge sharing and isolation.
\cite{lu2024visual} theoretically deduce two sufficient consistency conditions to strictly satisfy the orthogonality for prompt tuning, where the null space method~\cite{nullspace2021} is utilized to implement the conditions. 
Inspired by these methods, we also adopt the null space approach~\cite{nullspace2021} to construct the orthogonal projectors.
However, due to the higher-order recurrent state-space structure and the non-linear discretization operation in the SSM, the orthogonality constraints of the above methods are inapplicable to the Mamba architecture.
We aim to deduce the consistency conditions for the SSMs and design dedicated projectors to implement the orthogonal projection, as introduced in the next section.

{\subsection{Mamba-based Continual Learning}}
{Several recent studies have explored Mamba and State Space Models (SSMs) in different continual learning scenarios.
Mamba-FSCIL~\cite{Li2024MambaFSCILDA} employs a dual selective SSM projector and a class-sensitive selective scan for few-shot class-incremental learning, where the backbone is frozen during incremental sessions and old-class mean features are retained to facilitate adaptation to novel classes.
Learning Mamba as a Continual Learner~\cite{Zhao2024LearningMA} adopts a different meta-continual learning formulation, in which Mamba is meta-trained across multiple episodes as a sequence-prediction learner that incorporates streaming observations into its recurrent hidden state.
Inf-SSM~\cite{Lee_2026_CVPR} considers exemplar-free continual learning and introduces a geometry-aware method that regularizes the infinite-horizon state evolution encoded by the extended observability subspace of an SSM.
In contrast to these approaches, Mamba-CL continually adapts the SSM parameters of a pretrained visual Mamba backbone under the conventional exemplar-free class-incremental learning protocol and derives four parameter-level consistency conditions for the SSM.
Its null-space gradient projection explicitly protects the input-dependent discretization, state transition, input injection, and output readout processes involved in selective scanning.
Moreover, Mamba-CL does not require episodic meta-training, task-specific backbone expansion, or the retention of previous-task samples or feature embeddings, and its projection mechanism introduces no additional operations during inference.
These properties distinguish Mamba-CL from existing Mamba-based continual learning approaches and provide an effective way to continually adapt the core selective SSM modules while mitigating interference with previously acquired knowledge.}

\begin{figure*}[t]
	\centering
	\includegraphics[width=1.0\textwidth]{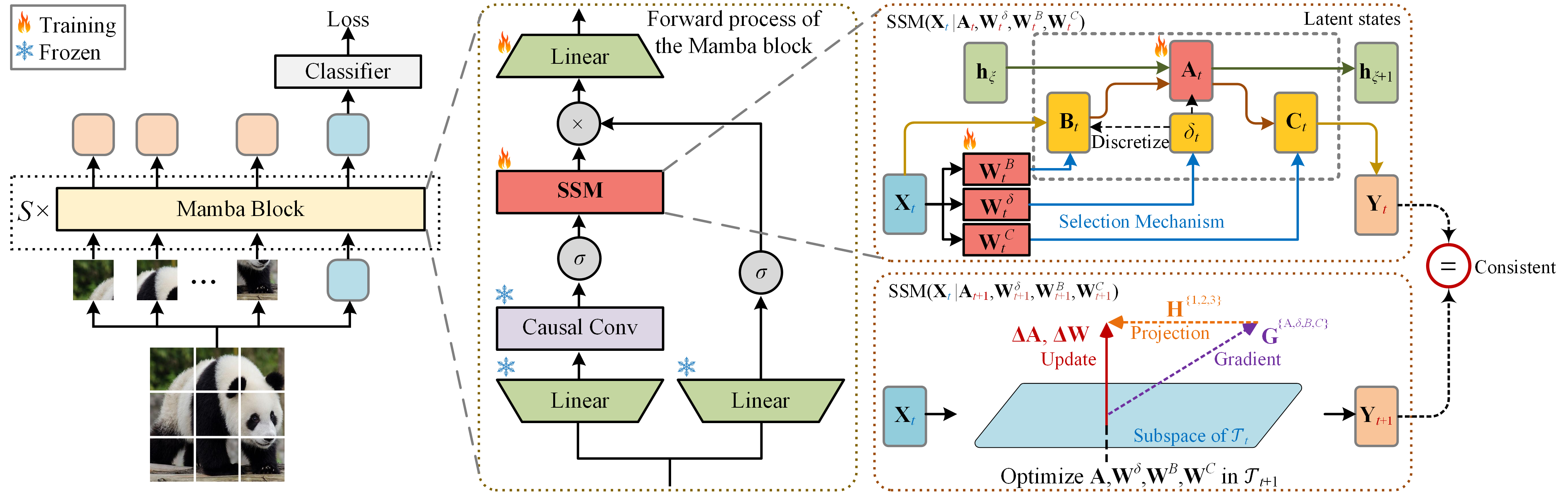}
	\caption{Illustration of the proposed Mamba-CL. The backbone contains $S$ Mamba blocks. For each Mamba block, we fine-tune the weights $ \vW^{\delta}, \vW^B, \vW^C$ and $\vA$ within the SSM (as well as the linear layer after the SSM). To minimize forgetting, we aim to utilize orthogonal projections to keep the output unchanged from SSM after training the $(t+1)$-th task.}
	\label{fig:overview}
\end{figure*}

\section{Preliminaries}

\noindent \textbf{Problem Definition.} In continual learning, there is a sequence of tasks with non-overlapping classes. We define the task sequence as $\mathcal{T} = \{\mathcal{T}_1, \cdots, \mathcal{T}_{T} \}$. $\mathcal{D}_t = \{\mathbf{x}_t^{<i>}, y_t^{<i>}\}_{i=1}^{|\mathcal{T}_t|}$ is the 
dataset associated with the $t$-th task of size $|\mathcal{T}_t|$, where $\mathbf{x}_t^{<i>} \in \mathbb{R}^{H \times W \times C}$ is one training image and $y_t^{<i>} \in \mathcal{Y}^t$ is the ground-truth label.
$\mathcal{Y}^t$ is the label space of the $t$-th task, and $\mathcal{Y}^t \cap \mathcal{Y}^{t'} = \varnothing$ for $t \neq t'$ in class-incremental learning. The objective of continual learning is to train a model $f(\cdot|\bm{\theta})$ with parameters $\bm{\theta}$ sequentially on $T$ tasks and perform well on all seen classes $\mathcal{Y}^1 \cup \cdots \cup \mathcal{Y}^T$.
Under the class-incremental learning protocol, the task identities are unknown during inference.

Inspired by existing fine-tuning-based CL methods \cite{l2pCVPR22, dualpromptECCV22, lu2024visual}, we utilize a pre-trained 
De-focus Mamba \cite{2024defocusmamba} as our backbone.
As illustrated in Fig.\ref{fig:overview}, Mamba-CL tunes the parameters within the SSM blocks and the linear projections after the SSM blocks via orthogonal projection-based optimization, while maintaining others frozen.


\subsection{The Forward Process of State Space Models}
In this section, we outline the forward process of the State Space Model (SSM) in Mamba.
Let $\mathbf{X} \in \mathbb{R}^{L \times D}$ denote an input sequence of SSM with token length $L$, where $D$ represents the dimension of each token. 
The SSM requires four input parameters, including an input-invariant parameter $\mathbf{A}$, and three input-conditioned parameters $\mathbf{B}$, $\mathbf{C}$ and $\bm{\delta}$ \cite{2023mamba}.
First, the input $\mathbf{X}$ undergoes three projection layers $s_B(\cdot)$, $s_C(\cdot)$ and $s_{\delta}(\cdot)$ to derive the above three input-conditioned parameters: 
\begin{equation}\label{eq: ssm_BCdelta_definition}
\left\{
\begin{aligned}
& \mathbf{B} = s_B(\mathbf{X}; \mathbf{W}^B) \in \mathbb{R}^{L \times N},\\
& \mathbf{C} = s_C(\mathbf{X}; \mathbf{W}^C) \in \mathbb{R}^{L \times N},\\
& \bm{\delta} = \tau_{\delta}(param + s_{\delta}(\mathbf{X}; \mathbf{W}^{\delta})) \in \mathbb{R}^{L \times 1},
\end{aligned}
\right.
\end{equation}
where $\mathbf{W}^B \in \mathbb{R}^{D \times N}$, $\mathbf{W}^C \in \mathbb{R}^{D \times N}$ and $\mathbf{W}^{\delta} \in \mathbb{R}^{D \times 1}$ denote the learnable weights in $s_B(\cdot)$, $s_C(\cdot)$ and $s_{\delta}(\cdot)$, respectively.
Here, $N$ denotes the dimension of each latent state in SSM.
$param$ is a bias term and $\tau_{\delta}(\cdot)$ denotes the softplus operation \cite{2023mamba} to avoid negative values in $\bm{\delta}$. 

After that, the SSM transforms the `continuous parameters' $(\mathbf{A}, \mathbf{B}, \bm{\delta})$ to `discrete parameters' $(\overline{\mathbf{A}}, \overline{\mathbf{B}})$ through the zero-order hold (ZOH) discretization rule:
\begin{equation}\label{eq: ssm_discreatization}
\left\{
\begin{aligned}
& \overline{\mathbf{A}} = \exp(\bm{\delta}\mathbf{A}) \in \mathbb{R}^{L \times D \times N},\\
& \overline{\mathbf{B}} = (\bm{\delta}\mathbf{A})^{-1} (\exp(\bm{\delta}\mathbf{A} - \mathbf{I})) \bm{\delta}\mathbf{B}\in \mathbb{R}^{L \times D \times N},
\end{aligned}
\right.
\end{equation}
where $\mathbf{I}$ denotes the identity matrix. 

After the discretization, the final output is derived through a global convolution with kernel $\overline{\mathbf{K}}$,
and the convolutional kernel is constructed on top of the above parameters $\overline{\mathbf{A}}$, $\overline{\mathbf{B}}$ and $\mathbf{C}$. The forward process is formulated as follows:
\begin{equation}\label{eq: ssm_build_conv}
    \overline{\mathbf{K}} = \left[
    \mathbf{C}\overline{\mathbf{B}},   \mathbf{C}\overline{\mathbf{A}}\overline{\mathbf{B}},\cdots,\mathbf{C}{\overline{\mathbf{A}}}^k\overline{\mathbf{B}},\cdots
    \right],
\end{equation}
\vspace{-6mm}
\begin{equation}\label{eq: ssm_global_convolution}
    \mathbf{Y} = SSM(\mathbf{X}|\mathbf{A}, \mathbf{W}^B, \mathbf{W}^C, \mathbf{W}^{\delta})= \mathbf{X} \ast \overline{\mathbf{K}},
\end{equation}
where $k$ denotes the index of the kernel tensor corresponding to the $k$-th latent state within the SSM.
$\mathbf{Y}$ denotes the output of the SSM model and `$\ast$' denotes the convolution operation.
The learnable parameters in the SSM including $\mathbf{A} \in \mathbb{R}^{D \times N}$ and the projection weights $\mathbf{W}^{\{B, C, \delta\}} \in \mathbb{R}^{D \times \{N,N,1\}}$. Finally, the output $\mathbf{Y}$ is fed into the next layer.

\subsection{Orthogonal Projection in Convolutional Layers}
For a convolutional layer $f_{conv}(\cdot | \mathbf{\Theta}_t)$ after training the $t$-th task, we define its unrolled convolutional matrix as $\mathbf{\Theta}_t \in \mathbb{R}^{D_{in} \times D_{out}}$, where $D_{in}$ denotes the number of pixels in each kernel and $D_{out}$ denotes the number of channels.
We define a flattened input from the $t$-th task as $\mathbf{X}_{t} \in \mathbb{R}^{N \times D_{in}}$ with $N$ patches.
The forward process is simply formulated as a linear operation, where $f_{conv}(\mathbf{X}_{t} | \mathbf{\Theta}_t) = \mathbf{X}_t\mathbf{\Theta}_t$.
The \emph{orthogonal projection} aims to keep the output unchanged from the convolutional weight after training $t+1$-th task to minimize forgetting, which is formulated as follows:
\begin{equation}\label{eq: ns_conv_formulation}
f_{conv}(\mathbf{X}_{t} | \mathbf{\Theta}_t) = 
f_{conv}(\mathbf{X}_{t} | \mathbf{\Theta}_{t+1}),
\end{equation}
where $\mathbf{\Theta}_{t+1} = \mathbf{\Theta}_t + {\Delta}\mathbf{\Theta}$. ${\Delta}\mathbf{\Theta}$ denotes the change of $\mathbf{\Theta}$ before and after learning the $t+1$-th task.
Then we derive the updating rule to achieve Eq.\eqref{eq: ns_conv_formulation}:
\begin{equation}\label{eq: ns_conv_condition}
	\mathbf{X}_t\mathbf{\Theta}_t = 
	\mathbf{X}_t(\mathbf{\Theta}_t + {\Delta}\mathbf{\Theta})
	\Rightarrow\mathbf{X}_t{\Delta}\mathbf{\Theta}=\mathbf{0}.
\end{equation}

The above term suggests that if the change of convolutional weight ${\Delta}\mathbf{\Theta}$ in the new task is orthogonal to the previous input feature space $\mathbf{X}_t$, the output of old task instances will remain unchanged.
Existing methods \cite{2021gpm, nullspace2021, 2023dualgpm} implement Eq.\eqref{eq: ns_conv_condition} by projecting the original gradient $\mathbf{G}^{{\Theta}}$ onto the orthogonal space of previous feature space through a projector $\mathbf{P} \in \mathbb{R}^{D_{in} \times D_{in}}$, where ${\Delta}\mathbf{\mathbf{\Theta}}=\mathbf{P}\mathbf{G}^{{\Theta}}$.

Similarly, for the SSM model with learnable parameters $\{\mathbf{A}, \mathbf{W}^{B}, \mathbf{W}^{C}, \mathbf{W}^{\delta}\}$, we also aim to remain the previous output unchanged while learning the new task:
\begin{equation}\label{eq: ns_ssm_formulation}
\begin{aligned}
&SSM(\mathbf{X}_t|\mathbf{A}_t, \mathbf{W}^B_t, \mathbf{W}^C_t, \mathbf{W}^{\delta}_t)=\\
&SSM(\mathbf{X}_t|\mathbf{A}_{t+1}, \mathbf{W}^B_{t+1}, \mathbf{W}^C_{t+1}, \mathbf{W}^{\delta}_{t+1}),
\end{aligned}
\end{equation}
where the subscripts $t$ and $t+1$ denote the parameters trained after the $t$-th and $t+1$-th tasks, respectively.
Due to the complex forward propagation in SSM, the simple consistency condition given by Eq.\eqref{eq: ns_conv_condition} is insufficient to achieve the above consistency term.
In the next section, we will deduce the conditions to achieve Eq.\eqref{eq: ns_ssm_formulation} in the SSM.

\section{Our Proposed Methodology}
{The overarching goal of this methodology is to derive conservative sufficient conditions that provide a tractable approximation to the general requirement of preserving SSM output consistency across tasks, thereby reducing the forgetting of previously acquired knowledge.}
This is achieved by analyzing the impact of parameter changes on the SSM's forward process and deducing constraints that maintain the consistency of SSM's output (\ie, Eq.\eqref{eq: ns_ssm_formulation}).
Specifically, we focus on four critical parameters in the Mamba model: the input-invariant parameter \( \mathbf{A} \), and the input-conditioned parameters \( \mathbf{W}^{B}, \mathbf{W}^{C} \) and \( \mathbf{W}^{\delta} \).
We use $\{\Delta\mathbf{A}, \Delta\mathbf{W}^B, \Delta\mathbf{W}^C, \Delta\mathbf{W}^{\delta}\}$ to denote the change of these parameters before and after learning the $(t+1)$-th task.
{By constraining the changes in these parameters to be orthogonal to the feature subspaces of previous tasks, we aim to reduce interference with previously learned knowledge.}

This section is structured as follows: First, we present the overall framework of Mamba-CL, outlining how orthogonal projections are integrated into the Mamba architecture for continual learning.
Subsequently, we analyze the consistency conditions required to satisfy Eq.\eqref{eq: ns_ssm_formulation} by breaking down the problem into three key components: the consistency of $\overline{\mathbf{A}}^{k}$, $\overline{\mathbf{B}}$, and $\mathbf{C}$.
For each component, we derive the sufficient conditions on the parameter changes $\Delta\mathbf{A}, \Delta\mathbf{W}^B, \Delta\mathbf{W}^C$, and $\Delta\mathbf{W}^{\delta}$.
Next, we propose an optimization strategy that leverages null-space projections to enforce these conditions during the training process.
Finally, we introduce a balance factor to relax the orthogonality constraints, allowing for a trade-off between stability (retaining old knowledge) and plasticity (learning new tasks).
By systematically addressing these challenges, we eventually provide a robust framework for continual learning with the Mamba model, ensuring that the model can adapt to new tasks while preserving the knowledge from previous ones.

\subsection{Overall Framework}
The overall framework of Mamba-CL is illustrated in \figurename\ref{fig:overview}.
Given an input image, it is first tokenized into a sequence of patches and processed through a series of Mamba blocks.
Each block contains a core SSM module with three linear layers and a causal convolutional layer.
During continual learning, only these SSM parameters and the linear projection layer after the SSM are fine-tuned, while the remaining components of the pre-trained Mamba foundation model remain frozen.
The final output is fed into task-specific classifiers for prediction, with cross-entropy loss applied to train the model.
During inference, the prediction of task-specific classifiers available so far will be concatenated for classification among all classes.

To optimize for a new task \( \calT_{t+1} \), the gradients of the SSM parameters (denoted as \(\vG^{\{A,B,C,\delta\}}\)) are projected onto subspaces orthogonal to the feature representations of previous tasks.
Specifically, we compute null-space projectors \( \vH^{\{1,2,3\}} \) of previous tasks.
These projectors enforce orthogonality constraints tailored to each parameter.
By updating parameters within these orthogonal subspaces, the outputs of SSM modules remain stable for previous tasks, effectively mitigating catastrophic forgetting.
The following subsections detail the theoretical derivation of these constraints and the optimization of consistency conditions.

\subsection{Analysis of the Consistency Conditions}
As shown in Fig.\ref{fig:overview}, the output is determined by the global convolution in Eq.\eqref{eq: ssm_global_convolution}.
Therefore, to achieve Eq.\eqref{eq: ns_ssm_formulation}, the previous output from Eq.\eqref{eq: ssm_global_convolution} should be unchanged before and after learning the new task.
We substitute Eq.\eqref{eq: ssm_global_convolution} into Eq.\eqref{eq: ns_ssm_formulation}:
\begin{equation}\label{eq: consistency_y}
    \mathbf{X}_t \ast \overline{\mathbf{K}}_t = 
    \mathbf{X}_t \ast \overline{\mathbf{K}}_{t+1}.
\end{equation}

\noindent To achieve the consistency condition formulated in the above term, we derive the following equation:
\begin{equation}\label{eq: consistency_k}
    \overline{\mathbf{K}}_t = \overline{\mathbf{K}}_{t+1}.
\end{equation}

\noindent 
Through the definition of $\overline{\mathbf{K}}$ in Eq.\eqref{eq: ssm_build_conv}, the above consistency term can be transformed as follows:
\begin{equation}\label{eq: consistency_ABC}
\begin{aligned}
&\left[
{\mathbf{C}}_t\overline{\mathbf{B}}_t,\cdots,\mathbf{C}_t{\overline{\mathbf{A}}}_t^k\overline{\mathbf{B}}_t,\cdots
\right]
=\\
&\left[
\mathbf{C}_{t+1}\overline{\mathbf{B}}_{t+1},\cdots,\mathbf{C}_{t+1}{\overline{\mathbf{A}}}_{t+1}^k\overline{\mathbf{B}}_{t+1},\cdots
\right].
\end{aligned}
\end{equation}

\noindent Since Eq.\eqref{eq: consistency_ABC} contains three variables, ($i.e.$, $\overline{\mathbf{A}}$, $\overline{\mathbf{B}}$ and $\mathbf{C}$), we analyze the sufficient conditions for which Eq.\eqref{eq: consistency_ABC} holds as a particular solution to the equation:
\begin{equation}\label{eq: consistency_ABC_2}
    \left\{
    \begin{aligned}
    &\overline{\mathbf{A}}_t^k = \overline{\mathbf{A}}_{t+1}^k,\\
    &\overline{\mathbf{B}}_t = \overline{\mathbf{B}}_{t+1},\\
    &\mathbf{C}_t = \mathbf{C}_{t+1}.
    \end{aligned}
    \right.
\end{equation}

Next, we separately analyze the conditions under which the above consistency equations can be satisfied. 

\subsubsection{Analysis of the condition to satisfy $\overline{\mathbf{A}}_t^k = \overline{\mathbf{A}}_{t+1}^k$}
According to the definition $\overline{\mathbf{A}}= \exp(\bm{\delta} \mathbf{A})$ in Eq.\eqref{eq: ssm_discreatization}, we aim to deduce the conditions for the two variables $\mathbf{A}$ and $\bm{\delta}$.
Based on the definition of $\overline{\mathbf{A}}$, we first transform the condition $\overline{\mathbf{A}}_t = \overline{\mathbf{A}}_{t+1}$ as follows:
\begin{equation}
   [\exp(\bm{\delta}_t \mathbf{A}_t)]^k = 
    [\exp(\bm{\delta}_{t+1} \mathbf{A}_{t+1})]^k.
\end{equation}

\noindent 
Since the exponential operation is element-wise, and given the monotonic increasing property of $[\exp(x)]^k$, the above term can be transformed as follows:
\begin{equation}\label{eq: condition_deltaA0}
    \bm{\delta}_t \mathbf{A}_t = 
    \bm{\delta}_{t+1} \mathbf{A}_{t+1}.
\end{equation}
{Since Eq.~\eqref{eq: condition_deltaA0} involves coupled changes in both $\bm{\delta}$ and $\mathbf{A}$, it may admit more general solutions in which their changes compensate for each other.}
{To obtain tractable parameter-wise constraints, we adopt the conservative sufficient condition $\bm{\delta}_t=\bm{\delta}_{t+1}$, which preserves the input-dependent discretization for previous-task inputs.}
{Under this condition, Eq.~\eqref{eq: condition_deltaA0} can be decomposed into the following two consistency conditions for $\bm{\delta}$ and $\mathbf{A}$:}
\begin{equation}\label{eq: condition_delta_A}
\left\{
\begin{aligned}
& \bm{\delta}_t = \bm{\delta}_{t+1}, \\
& \bm{\delta}_{t} \mathbf{A}_{t} = \bm{\delta}_{t} \mathbf{A}_{t+1} = \bm{\delta}_{t} (\mathbf{A}_{t} + \Delta\mathbf{A})\ 
\Rightarrow
\bm{\delta}_{t} \Delta\mathbf{A} = \mathbf{0}.\\
\end{aligned}
\right.
\end{equation}
{This decomposition provides a sufficient condition rather than a necessary-and-sufficient characterization, because more general consistency-preserving updates may allow changes in $\bm{\delta}$ and $\mathbf{A}$ to compensate for each other.}

To achieve $\bm{\delta}_t = \bm{\delta}_{t+1}$, given the definition $\bm{\delta} = \tau_{\delta}(param + s_{\delta}(\mathbf{X}; \mathbf{W}^{\delta}))$ in Eq.\eqref{eq: ssm_BCdelta_definition}, we aim to obtain the consistency conditions for $\mathbf{W}^{\delta}$.
Since the softplus function $\tau_{\delta}$ is monotonically increasing, and given $s_{\delta}(\mathbf{X}; \mathbf{W}^{\delta}) = \mathbf{X}\mathbf{W}^{\delta}$, the upper formula in Eq.\eqref{eq: condition_delta_A} can be simplified as follows:
\begin{equation}
\mathbf{X}_t\mathbf{W}^{\delta}_t = \mathbf{X}_t\mathbf{W}^{\delta}_{t+1} \Rightarrow
\mathbf{X}_t\mathbf{W}^{\delta}_t = \mathbf{X}_t(\mathbf{W}^{\delta}_t + \Delta\mathbf{W}^{\delta}).
\end{equation}

\noindent In this way, we derive the consistency condition for the trainable parameter $\Delta\mathbf{W}^{\delta}$:
\begin{equation}\label{eq: condition_delta}
\mathbf{X}_t\Delta\mathbf{W}^{\delta} = \mathbf{0}.    
\end{equation}

By combining Eq.\eqref{eq: condition_delta_A} and Eq.\eqref{eq: condition_delta}, we obtain two sufficient conditions of $\Delta\mathbf{A}$ and $\Delta\mathbf{W}^{\delta}$ for maintaining the consistency term $\overline{\mathbf{A}}^k_t = \overline{\mathbf{A}}^k_{t+1}$:
\begin{equation}\label{eq: condition_delta_A_final}
\left\{
\begin{aligned}
& \bm{\delta}_t \Delta\mathbf{A} = \mathbf{0},\\
&\mathbf{X}_t\Delta\mathbf{W}^{\delta} = \mathbf{0}.
\end{aligned}\right.
\end{equation}

\subsubsection{Analysis of the condition to satisfy $\overline{\mathbf{B}}_t = \overline{\mathbf{B}}_{t+1}$}
According to the definition $\overline{\mathbf{B}} = (\bm{\delta}\mathbf{A})^{-1} (\exp(\bm{\delta}\mathbf{A} - \mathbf{I})) \bm{\delta}\mathbf{B}$ in Eq.\eqref{eq: ssm_discreatization}, we formulate the consistency term \( \overline{\mathbf{B}}_t = \overline{\mathbf{B}}_{t+1} \) in Eq.\eqref{eq: consistency_ABC_2} as follows:
\begin{equation}
\begin{aligned}
    &(\bm{\delta}_t\mathbf{A}_t)^{-1} (\exp(\bm{\delta}_t\mathbf{A}_t - \mathbf{I})) \bm{\delta}_t\mathbf{B}_t =\\ 
    &(\bm{\delta}_{t+1}\mathbf{A}_{t+1})^{-1} (\exp(\bm{\delta}_{t+1}\mathbf{A}_{t+1} - \mathbf{I})) \bm{\delta}_{t+1}\mathbf{B}_{t+1}.
\end{aligned}
\end{equation}
Considering the deduced condition $\bm{\delta}_t \mathbf{A}_t = \bm{\delta}_{t+1} \mathbf{A}_{t+1}$ given by Eq.\eqref{eq: condition_deltaA0}, the above equation can be simplified as follows:
\begin{equation}
    \bm{\delta}_t \mathbf{B}_t = 
    \bm{\delta}_{t+1} \mathbf{B}_{t+1}.
\end{equation}

Given that $\mathbf{B} = \mathbf{X}\mathbf{W}^B$ is the output of the projection layer $s_B(\cdot)$, the above equation can be expanded as:
\begin{equation}
\bm{\delta}_t \mathbf{X}_t \mathbf{W}^B_t = 
\bm{\delta}_{t+1} \mathbf{X}_{t} (\mathbf{W}^B_{t} + \Delta\mathbf{W}^B).
\end{equation}

\noindent Considering the condition $\bm{\delta}_{t} = \bm{\delta}_{t+1}$ given by Eq.\eqref{eq: condition_delta_A} for parameter $\bm{\delta}$, we can derive the consistency condition for $\Delta\mathbf{W}^B$ by expanding the above equation:
\begin{equation}\label{eq: condition_wb}
    \bm{\delta}_t\mathbf{X}_t\Delta\mathbf{W}^B = \mathbf{0}.
\end{equation}

\subsubsection{Analysis of the condition to satisfy $\mathbf{C}_t = \mathbf{C}_{t+1}$}
Given the definition of $\mathbf{C}$ in Eq.\eqref{eq: ssm_BCdelta_definition}, where $\mathbf{C}=s_C(\mathbf{X}; \mathbf{W}^C)$, we reformulate $\mathbf{C}_t = \mathbf{C}_{t+1}$ as follows:
\begin{equation}
    s_C(\mathbf{X}_t; \mathbf{W}^C_t) = s_C(\mathbf{X}_t; \mathbf{W}^C_{t+1}).
\end{equation}

\noindent Given the linear operation in the projection layer $s_C(\cdot)$, where $s_C(\mathbf{X}; \mathbf{W}^C)=\mathbf{X}\mathbf{W}^C$, the consistency condition for $\Delta\mathbf{W}^C$ can be similarly derived:
\begin{equation}\label{eq: condition_w0}
	\mathbf{X}_t \mathbf{W}^C_t = \mathbf{X}_t \mathbf{W}^C_{t+1} = \mathbf{X}_t (\mathbf{W}^C_t + \Delta\mathbf{W}^C),
\end{equation}
which is further simplified as:
\begin{equation}\label{eq: condition_wc}
\mathbf{X}_t\Delta\mathbf{W}^C=\mathbf{0}.
\end{equation}

By combining Eq.\eqref{eq: condition_delta_A_final}, Eq.\eqref{eq: condition_wb} and Eq.\eqref{eq: condition_wc}, we finally derive four sufficient consistency conditions to satisfy Eq.\eqref{eq: ns_ssm_formulation} for the four updating parameters $\{\mathbf{A}, \mathbf{W}^B, \mathbf{W}^C, \mathbf{W}^{\delta}\}$:
\begin{equation}\label{eq: four_conditions}
\left\{
\begin{aligned}
& \bm{\delta}_t\Delta\mathbf{A} = \mathbf{0},\\
& \mathbf{X}_t\Delta\mathbf{W}^{\delta} = \mathbf{0},\\
&\bm{\delta}_t\mathbf{X}_t\Delta\mathbf{W}^B = \mathbf{0},\\
&\mathbf{X}_t\Delta\mathbf{W}^C=\mathbf{0}.
\end{aligned}\right.
\end{equation}

{The four conditions in Eq.~\eqref{eq: four_conditions} follow the same principle: parameter updates for a new task should have limited influence when the model processes previous-task inputs. Specifically, $\mathbf{W}^{\delta}$ and $\mathbf{A}$ control the input-dependent update scale and hidden-state propagation, while $\mathbf{W}^{B}$ and $\mathbf{W}^{C}$ control how input information is written into the hidden state and read out as the output, respectively. Therefore, these conditions jointly aim to preserve the state-update and output-generation process of previous-task samples, while leaving orthogonal directions available for learning new tasks.}
{They are sufficient under the adopted decoupled formulation and provide a tractable approximation to the broader set of consistency-preserving updates.}
Therefore, during training in the new task, we aim to keep that the conditions in Eq.\eqref{eq: four_conditions} hold while optimizing the four parameters \( \{\mathbf{A}, \mathbf{W}^B, \mathbf{W}^C, \mathbf{W}^{\delta}\} \).
In the next section, we will describe the optimization process for updating the parameters under the above conditions in detail.

\subsection{Optimization of Consistency Conditions}\label{sec: null_space_optimizaion}
To jointly optimize these four conditions shown in Eq.\eqref{eq: four_conditions}, we aim to solve the update of parameters $\{\Delta\mathbf{A}, \Delta\mathbf{W}^B, \Delta\mathbf{W}^C, \Delta\mathbf{W}^{\delta}\}$ to satisfy all conditions concurrently.
The second and fourth conditions show that the optimization should make $\Delta\mathbf{W}^{\delta}$ and $\Delta\mathbf{W}^C$ orthogonal to the feature space spanned by $\mathbf{X}_t$. 
Subsequently, the optimization should guarantee that $\Delta\mathbf{A}$ and $\Delta\mathbf{W}^B$ are orthogonal to the space spanned by $\bm{\delta}_t$ and $\bm{\delta}_t\mathbf{X}_t$, respectively.
In order to solve the conditions, we utilize $\mathbf{G}^{\{A, B, C, \delta\}}$ to represent the gradient of the parameters generated by the optimizer. We aim to find a group of projectors $\mathbf{P}^{\{A, B, C, \delta\}}$ that can project these gradients onto the orthogonal space of the aforementioned feature space to satisfy Eq.\eqref{eq: four_conditions}:
\begin{equation}
\left\{
\begin{aligned}
&\Delta\mathbf{A} = \mathbf{P}^A\mathbf{G}^A,\\
&\Delta\mathbf{W}^{\{B,C,\delta\}} = \mathbf{P}^{\{B,C,\delta\}}\mathbf{G}^{\{B,C,\delta\}}.\\
\end{aligned}\right.
\end{equation}

\begin{algorithm}[!ht]
	\caption{Null Space Optimization for Mamba-CL.}
	\label{alg:1}
	\begin{algorithmic}[1]
		\REQUIRE Datasets \( \mathcal{D}_t = \{(\mathcal{X}_t^{<i>}, y_t^{<i>})\}_{i=1}^{|\mathcal{T}_t|} \) for task \( \mathcal{T}_{t} \in \{\mathcal{T}_1, \mathcal{T}_2, \cdots \} \), the Mamba model \( f(\cdot) \) with learnable parameters $\{\mathbf{A}_{t-1}, \mathbf{W}^B_{t-1}, \mathbf{W}^C_{t-1}, \mathbf{W}^{\delta}_{t-1}\}$, 
		the uncentered covariance matrices $\{\mathbf{Q}_1, \mathbf{Q}_2, \mathbf{Q}_3\}$, the
		projection matrices $\{\mathbf{H}_1, \mathbf{H}_2, \mathbf{H}_3\}$, learning rate $\gamma$.
		\ENSURE Optimized parameters $\{\mathbf{A}_{t}, \mathbf{W}^B_{t}, \mathbf{W}^C_{t}, \mathbf{W}^{\delta}_{t}\}$.
		\STATE \textbf{Initialization:}
		$\{\mathbf{Q}_1, \mathbf{Q}_2, \mathbf{Q}_3\} = \{\mathbf{0}, \mathbf{0}, \mathbf{0}\}$;
		\FOR{task \( \mathcal{T}_t \in \{\mathcal{T}_1, \mathcal{T}_2, \cdots\} \)}
		\REPEAT
		\STATE Sample a mini-batch \( \boldsymbol{\mathcal{X}}_t, \boldsymbol{y}_t \sim \mathcal{D}_{t} \);
		\STATE Get \( \boldsymbol{\hat{y}}_t \gets f(\boldsymbol{\mathcal{X}}_t | \mathbf{A}_{t-1}, \mathbf{W}^B_{t-1}, \mathbf{W}^C_{t-1}, \mathbf{W}^{\delta}_{t-1})\);
		\STATE Compute loss \( \mathcal{L} \gets \mathrm{CrossEntropy}(\boldsymbol{\hat{y}}_t, \boldsymbol{y}_t) \);
		\STATE Backward propagation;
		\STATE Get the gradients $\mathbf{G}^{A}, \mathbf{G}^{B}, \mathbf{G}^{C}, \mathbf{G}^{\delta}$;
		\IF{\( t>1 \)}
		\STATE Update $\mathbf{W}^{\delta}$ by $\mathbf{W}^{\delta}_{t-1} \gets \mathbf{W}^{\delta}_{t-1} - \gamma\mathbf{H}_1\mathbf{G}^{\delta}$;\\
		\STATE Update $\mathbf{W}^{C}$ by $\mathbf{W}^{C}_{t-1} \gets \mathbf{W}^{C}_{t-1} - \gamma\mathbf{H}_1\mathbf{G}^{C}$;\\
		\STATE Update $\mathbf{A}$ by $\mathbf{A}_{t-1} \gets \mathbf{A}_{t-1} - \gamma\mathbf{H}_2\mathbf{G}^{A}$;\\
		\STATE Update $\mathbf{W}^B$ by $\mathbf{W}^B_{t-1} \gets \mathbf{W}^B_{t-1} - \gamma\mathbf{H}_3\mathbf{G}^{B}$;\\
		\ELSE
		\STATE Update $\mathbf{W}^{\delta}$ by $\mathbf{W}^{\delta}_{t-1} \gets \mathbf{W}^{\delta}_{t-1} - \gamma\mathbf{G}^{\delta}$;\\
		\STATE Update $\mathbf{W}^{C}$ by $\mathbf{W}^{C}_{t-1} \gets \mathbf{W}^{C}_{t-1} - \gamma\mathbf{G}^{C}$;\\
		\STATE Update $\mathbf{A}$ by $\mathbf{A}_{t-1} \gets \mathbf{A}_{t-1} - \gamma\mathbf{G}^{A}$;\\
		\STATE Update $\mathbf{W}^B$ by $\mathbf{W}^B_{t-1} \gets \mathbf{W}^B_{t-1} - \gamma\mathbf{G}^{B}$;\\
		\ENDIF
		\UNTIL{convergence}	
		\STATE Get optimized parameters $\{\mathbf{A}_{t}, \mathbf{W}^B_{t}, \mathbf{W}^C_{t}, \mathbf{W}^{\delta}_{t}\}$=
		$\{\mathbf{A}_{t-1}, \mathbf{W}^B_{t-1}, \mathbf{W}^C_{t-1}, \mathbf{W}^{\delta}_{t-1}\}$;
		\STATE Initialize three temporary matrices \( \vJ_1 = [~] \), \( \vJ_2 = [~] \) and \( \vJ_3 = [~] \);
		\FOR{\( \mathcal{X}_t^{<i>} \in \mathcal{D}_t \)}
		\STATE 
		Get the feature matrices \( (\mathbf{X}_t)^{<i>} \), \( (\bm{\delta}_t)^{<i>} \) and \( (\bm{\delta}_t\mathbf{X}_t)^{<i>} \) through the forward propagation;
		\STATE Update \( \vJ_1 \), \( \vJ_2 \) and \( \vJ_3 \) by concatenating \( (\mathbf{X}_t)^{<i>} \), \( (\bm{\delta}_t)^{<i>} \) and \( (\bm{\delta}_t\mathbf{X}_t)^{<i>} \), respectively;
		\ENDFOR
		\STATE Update uncentered covariance matrices \( \vQ_1 \gets \vQ_1 + \vJ_1^{\top} \vJ_1 \),  \( \vQ_2 \gets \vQ_2 + \vJ_2^{\top} \vJ_2 \) and \( \vQ_3 \gets \vQ_3 + \vJ_3^{\top} \vJ_3 \);
		\STATE Compute the projection matrices \(\{ \mathbf{H}_1, \mathbf{H}_2, \mathbf{H}_3 \}\) by SVD on \(\{\vQ_1, \vQ_2, \vQ_3\}\).
		\ENDFOR
	\end{algorithmic}
\end{algorithm}

Inspired by the null-space optimization methods \cite{2021gpm, lu2024visual, nullspace2021}, the bases of the projection matrices $\mathbf{P}^{\delta}$ and $\mathbf{P}^{C}$ should reside in the null space of $\mathbf{X}_t$.
Meanwhile, the bases of $\mathbf{P}^{A}$ and $\mathbf{P}^B$ should lie in the null space of $\bm{\delta}_t$ and $\bm{\delta}_t\mathbf{X}_t$, respectively.
To obtain the null space bases, after training the $t$-th task, we extract the feature matrices $\overline{\mathbf{X}}_t$, $\overline{\bm{\delta}}_t$, $\overline{\bm{\delta}_t\mathbf{X}}_t$ of all images from the $t$-th task.
For example, $\overline{\mathbf{X}}_t = [\mathbf{X}_t^1,\cdots, \mathbf{X}_t^m \cdots, \mathbf{X}_t^M] \in \mathbf{R}^{ML \times D}$ is built by the entire $t$-th dataset.
$\mathbf{X}_t^m$ is the feature of the $m$-th data point, and $M$ is the number of data points.
Then we derive the uncentered covariance matrices of the feature matrices, where $\mathbf{Q}_1 = \overline{\mathbf{X}}_t^{\top}\overline{\mathbf{X}}_t$, 
$\mathbf{Q}_2 = \overline{\bm{\delta}}_t^{\top}\overline{\bm{\delta}}_t$, and
$\mathbf{Q}_3 = \overline{\bm{\delta}_t\mathbf{X}}_t^{\top}\overline{\bm{\delta}_t\mathbf{X}}_t$.
Next, the null space bases $\mathbf{U}_{1,0}$, $\mathbf{U}_{2,0}$ and $\mathbf{U}_{3,0}$ are obtained from the right singular vectors corresponding to the zero singular values through singular value decomposition (SVD) of $\{\mathbf{Q}_1, \mathbf{Q}_2, \mathbf{Q}_3\}$.
Each group of null space bases constructs a subspace orthogonal to the corresponding feature subspace (\ie, ${\mathbf{X}}_t$, ${\bm{\delta}}_t$ and ${\bm{\delta}_t\mathbf{X}}_t$).

Note that since zero singular values do not always exist, we use an adaptive method to select the singular values close to zero as approximated zero singular values~\cite{nullspace2021}.
Based on the observation that the singular values in descending order form an "L" shape \cite{lu2024visual}, we first find the index of the corner point by calculating the maximum second derivative of the points: $R = J - \arg\max_{j} \{ \lambda_{j-1} - 2 \lambda_{j} + \lambda_{j+1} \}_{j=2}^{J-1}$, where $J$, $R$ and $\lambda_{j}$ denote the total number of singular values, the number of approximated zero singular values and the $j$-th singular value, respectively.
Those right singular values corresponding to the $R$ smallest singular values (\ie, the singular values below the corner point) obtained by SVD are selected as the null space bases.
In the absence of an exact null space, the approximate null space corresponding to singular values close to zero is a minimum-norm solution~\cite{golub2013matrix}, which means that the obtained subspace provides the most accurate approximation.
As a result, we derive three projectors $\mathbf{H}_{1} =\mathbf{U}_{1,0}\mathbf{U}_{1,0}^{\top}$, $\mathbf{H}_2=\mathbf{U}_{2,0}\mathbf{U}_{2,0}^{\top}$, and $\mathbf{H}_3=\mathbf{U}_{3,0}\mathbf{U}_{3,0}^{\top}$ to enable that the parameter updates satisfy the conditions in Eq.\eqref{eq: four_conditions}:
\begin{equation}\label{eq: final_projection}
\left\{
\begin{aligned}
& \Delta\mathbf{W}^{\delta} = \mathbf{H}_{1} \mathbf{G}^{\delta},\\
& \Delta\mathbf{W}^C = \mathbf{H}_{1} \mathbf{G}^C,\\
& \Delta\mathbf{A} = \mathbf{H}_2 \mathbf{G}^A,\\
& \Delta\mathbf{W}^B = \mathbf{H}_3 \mathbf{G}^B.
\end{aligned}\right.
\end{equation}

To sum up, during training, we use Eq.\eqref{eq: final_projection} to perform orthogonal projections for parameter updating. 
The whole training process is outlined in Algorithm \ref{alg:1}.
To further balance the stability and plasticity of the model, we introduce a balance hyper-parameter $\eta \in [0,1]$ to relax the orthogonality constraints.
For example, for $\Delta\mathbf{W}^{\delta}$, we combine the null space projection $\mathbf{H}_{1}$ and the identity matrix $\mathbf{I}$ to derive a new projector $\overline{\mathbf{H}}_{1}$ for enhancing plasticity: 
$\overline{\mathbf{H}}_{1} = \eta \mathbf{H}_{1} + (1-\eta)\mathbf{I}$.
$\eta$ represents the strictness degree of the orthogonality which should be close to 1.
{When $\eta=1$, $\overline{\mathbf{H}}=\mathbf{H}$, so Mamba-CL reduces to strict null-space projection, which is equivalent to removing the relaxation controlled by $\eta$.}
After that, the updating rule of $\mathbf{W}^{\delta}$ is given by:
$\Delta\mathbf{W}^{\delta} = \overline{\mathbf{H}}_{1} \mathbf{G}^{\delta}$.
Such a strategy can be applied to other parameters similarly to improve the model's plasticity.

\section{Experiments}
\subsection{Experimental Setups}
\label{sec:exp_setup}

\textbf{Benchmark Datasets:}
We conduct experiments under the class-incremental learning protocol, where the classes in each task are disjoint, and task identity is unknown during inference.
Four class-incremental benchmarks across three widely used datasets are mainly adopted: 10-split and 20-split ImageNet-R \cite{dualpromptECCV22}, 10-split CIFAR-100 \cite{cifar100} and 10-split DomainNet \cite{domainet, 2023esn}.
We follow \cite{dualpromptECCV22} to randomly split the 200 classes in ImageNet-R into 10 and 20 tasks, forming the 10-split and 20-split ImageNet-R benchmarks, respectively, aiming to evaluate the ability to handle different numbers of tasks.
For the CIFAR-100 dataset, the total 100 classes are randomly split into 10 tasks.
Note that the DomainNet~\cite{domainet} was originally proposed for domain-incremental learning.
However, we follow the same dataset protocol adopted in \cite{2023esn} and \cite{cpromptCVPR24} to select the top 200 classes with the most images from the original DomainNet, and randomly split them into 10 tasks with 20 classes per task to construct a cross-domain class-incremental benchmark.
Furthermore, the above three datasets are also split into 50 or 100 tasks to evaluate the long-sequence continual learning performance in our experiments.

\textbf{Model Setup:}
As we investigated, only the De-focus Mamba-Large~\cite{2024defocusmamba} provides the pre-training weights on the ImageNet-21k dataset~\cite{imagenet-21k}, while other variants such as De-focus Mamba-Base~\cite{2024defocusmamba}, MambaVision~\cite{hatamizadeh2024mambavision}, VMamba~\cite{2024vmamba} and Vim~\cite{2024visionmamba} only have the pre-training weights on the ImageNet-1k dataset.
To compare with existing ViT-based methods which are pre-trained on ImageNet-21k, we mainly employ the De-focus Mamba-Large which is also pre-trained on ImageNet-21k as the backbone in our experiments.
In order to further verify the generalizability of our approach across various Mamba variants and compare with existing ViT-based models under similar backbones as well as the same pre-training weights, we also experiment on MambaVision~\cite{hatamizadeh2024mambavision}, Vim~\cite{2024visionmamba} and VMamba~\cite{2024vmamba} which are pre-trained on the ImageNet-1k dataset, whose results are shown in Section~\ref{sec:variants}.

There are 48 Mamba blocks in the backbone and the SSM in every Mamba block is fine-tuned with the proposed null-space projection.
In addition to the SSM block, the liner projection layer after the SSM is also trained with null-space projection for a better adaptation to downstream tasks.
The null-space projection matrix for the linear layer after the SSM is computed in the same way as that in linear/convolutional layers~\cite{nullspace2021}.
The classifiers are trained for each task independently.
{Previously learned classifier weights are retained unchanged, while only the weights associated with newly introduced classes are optimized. The newly introduced classifier weights have no previous values to preserve and require sufficient flexibility to learn discriminative directions for new classes. Applying null-space constraints constructed from previous-task features to these weights is therefore not well suited to our classifier design.}
During inference, all classifiers from previous tasks are concatenated to make predictions for all available classes.

\textbf{Implementation Details:}
The images fed into the models are resized to \( 192 \times 192 \) pixels and augmented by AutoAugment \cite{cubuk2019autoaugment} during training.
We use the Adam optimizer \cite{kingma2014adam} with \( \beta_1 = 0.9 \), \( \beta_2 = 0.999 \) and a weight decay of \( 5 \times 10^{-5} \) to train the backbone for 50 epochs with an initial learning rate of 0.0002 and a batch size of 200.
The learning rate is scaled by a factor of 0.1 at the 25-th and 40-th epochs.
For the classifier, we use a large learning rate of 0.01 to promote the adaptation to downstream tasks.
Our training loss only involves the cross-entropy loss for classification.
As introduced in the Optimization of Consistency Conditions section, we adopt a balanced hyper-parameter \( \eta \) for the trade-off between stability and plasticity in the null-space projections.
\( \eta \) is set to 0.90 for 10-split ImageNet-R and 0.95 for other benchmarks.
We implement our approach in PyTorch \cite{paszke2019pytorch} with the timm library \cite{rw2019timm}.
The experiments are performed on a server with 128 GB RAM and four NVIDIA RTX 4090 GPUs.


\begin{table*}[t]
	\centering
	\setlength{\tabcolsep}{0.5pt}
	\caption{Comparison with existing methods pre-trained on the ImageNet-21K. The results marked with \tdr~and * are implemented by \cite{cpromptCVPR24} and us, respectively, due to a lack of official results. The highest accuracies are in bold, and the second-highest accuracies are underlined. The value after \( \pm \) indicates the standard deviation.}
	\resizebox{\linewidth}{!}{
		\begin{tabular}{@{}llcccccccc@{}}
			\toprule
			\multirow{2}{*}{Method} & \multirow{2}{*}{Venue} & \multicolumn{2}{c}{10-split ImageNet-R}           & \multicolumn{2}{c}{20-split ImageNet-R}           & \multicolumn{2}{c}{10-split CIFAR-100}            & \multicolumn{2}{c}{10-split DomainNet}              \\ \cmidrule(l){3-10} 
			&                        & Acc.\uar                    & Forgetting\dar      & Acc.\uar                    & Forgetting\dar      & Acc.\uar                    & Forgetting\dar      & Acc.\uar                    & Forgetting\dar        \\ \midrule
			L2P~\cite{l2pCVPR22}                & CVPR'22                & 61.57\pms{0.66}             & 9.73\pms{0.47}      & 59.38\pms{0.50}             & 5.89\pms{0.36}      & 83.83\pms{0.04}             & 7.63\pms{0.30}      & ~\,81.17\pms{0.83}\tdr      & ~~8.98\pms{1.25}      \\
			DualPrompt~\cite{dualpromptECCV22}  & ECCV'22                & 68.13\pms{0.49}             & 4.68\pms{0.20}      & 63.21\pms{0.49}             & 5.28\pms{0.45}      & 86.51\pms{0.33}             & 5.16\pms{0.09}      & ~\,81.70\pms{0.78}\tdr      & ~~8.04\pms{0.31}      \\
			CODA-Prompt~\cite{codaCVPR23}       & CVPR'23                & 75.45\pms{0.56}             & 1.64\pms{0.10}      & 72.37\pms{1.19}             & 0.96\pms{0.15}      & 86.25\pms{0.74}             & 1.67\pms{0.26}      & ~\,80.04\pms{0.79}\tdr      & 10.16\pms{0.35}       \\
			LAE~\cite{laeICCV23}                & ICCV'23                & 72.66\pms{0.63}             & -                   & 69.67\pms{0.86}             & -                   & 85.59\pms{0.46}             & -                   & -                           & -                     \\
			LGCL~\cite{2023lgcl}    			& ICCV'23                & 69.46\pms{0.04}             & 4.20\pms{0.06}      & -                           & -                   & 87.23\pms{0.21}             & 5.10\pms{0.15}      & -                           & -                     \\
			C-LN~\cite{2023c-ln}              & ICCVW'23               & 76.36\pms{0.51}             & 8.31\pms{1.28}      & 71.72\pms{0.47}             & 5.42\pms{0.28}      & 86.95\pms{0.37}             & 6.98\pms{0.43}      & -                           & -                     \\
			ESN~\cite{2023esn}                & AAAI'23                & 71.07\pms{0.29}             & 4.99\pms{0.49}      & 64.77\pms{0.71}             & 6.65\pms{1.24}      & 86.34\pms{0.52}             & 4.76\pms{0.14}      & ~\,79.22\pms{2.04}\tdr      & 10.62\pms{2.12}       \\
			C-ADA~\cite{PromptLearning2024Gao}						& ECCV'24				 & 76.66\tsb{\qquad~~}		   & -					 & 73.47\tsb{\qquad~~}		   & - 					 & 87.18\tsb{\qquad~~}		   & -					 & - 						   & -					   \\
			OS-Prompt++~\cite{OneStagePromptBased2024Kim}			& ECCV'24				 & 75.67\pms{0.40}		       & 1.27 \pms{0.10}	 & 73.77\pms{0.19}		  	   & 0.79\pms{0.07}		 & 86.68\pms{0.67}			   & 1.18\pms{0.21}  	 & - 						   & -					   \\
			EvoPrompt~\cite{evopromptAAAI24}    & AAAI'24                & 76.83\pms{0.08}             & 2.78\pms{0.06}      & 74.41\pms{0.23}             & 2.56\pms{0.22}      & 87.97\pms{0.30}             & 2.60\pms{0.42}      & ~\,79.50\pms{0.29}\str      & ~~3.81\pms{0.36}      \\
			PGP~\cite{pgpICLR24}                & ICLR'24                & 69.34\pms{0.05}             & 4.53\pms{0.04}      & -                           & -                   & 86.92\pms{0.05}             & 5.35\pms{0.19}      & ~\,80.41\pms{0.25}\str      & ~~8.39\pms{0.18}      \\
			OVOR-Deep~\cite{2024ovorprompt}         & ICLR'24                & 76.11\pms{0.21}             & 7.16\pms{0.34}      & 73.85\pms{0.29}             & 6.80\pms{0.65}      & 85.99\pms{0.89}             & 6.42\pms{2.03}      & ~\,79.61\pms{0.86}\str      & ~~4.77\pms{0.94}      \\
			ConvPrompt~\cite{2024ConvPrompt}  & CVPR'24                & 77.86\pms{0.25}             & 4.33\pms{0.24}      & 75.10\pms{0.39}             & 4.10\pms{0.29}      & 88.87\pms{0.33}             & 4.75\pms{0.15}      & ~\,79.47\pms{0.35}\str      & ~~6.49\pms{0.43}      \\
			InfLoRA~\cite{infloraCVPR24}        & CVPR'24                & 75.65\pms{0.14}             & 5.73\pms{0.44}      & 71.01\pms{0.45}             & 6.83\pms{0.44}      & 87.06\pms{0.25}             & 6.22\pms{0.39}      & ~\,81.45\pms{0.68}\str      & ~~5.35\pms{0.52}      \\
			EASE~\cite{easeCVPR24}              & CVPR'24                & 76.17\tsb{\qquad~~}        & 7.82\tsb{\qquad~~} & 73.27\tsb{\qquad~~}        & 8.51\tsb{\qquad~~} & 87.76\tsb{\qquad~~}        & 5.94\tsb{\qquad~~} & 78.89\str\tsb{\qquad}      & ~~7.89\tsb{\qquad~~} \\
			CPrompt~\cite{cpromptCVPR24}        & CVPR'24                & 77.14\pms{0.11}             & 5.97\pms{0.68}      & 74.79\pms{0.28}             & 7.34\pms{0.65}      & 87.82\pms{0.21}             & 5.06\pms{0.50}      & \underline{82.97}\pms{0.34} & ~~7.45\pms{0.93}      \\
				{SD-LoRA~\cite{SDLoRAScalable2025Wu}}
				& {ICLR'25}
				& {77.34\pms{0.35}} & {5.85\pms{0.24}}
				& {75.26\pms{0.37}} & {6.24\pms{0.28}}
				& {88.01\pms{0.31}} & {4.93\pms{0.22}}
				& {~\,81.01\pms{0.42}} & {~~7.03\pms{0.35}} \\
				{CPrompt-KAC~\cite{KACKolmogorovArnold2025Hu}}
				& {CVPR'25}
				& {78.07\tsb{\qquad~~}} & {5.55\tsb{\qquad~~}}
				& {75.73\tsb{\qquad~~}} & {6.04\tsb{\qquad~~}}
				& {87.19\tsb{\qquad~~}} & {5.22\tsb{\qquad~~}}
				& {~\,80.46\tsb{\qquad~~}} & {~~7.53\tsb{\qquad~~}} \\
				{BiLoRA~\cite{BiLoRAAlmostOrthogonal2025Zhu}}
				& {CVPR'25}
				& {77.95\pms{0.14}} & {5.37\pms{0.28}}
				& {72.41\pms{0.76}} & {7.58\pms{0.42}}
				& {87.46\pms{0.76}} & {5.07\pms{0.31}}
				& {~\,76.56\pms{0.41}} & {~~8.24\pms{0.45}} \\
				{LoRA-DRS~\cite{LoRASubtraction2025Liu}}
				& {CVPR'25}
				& {75.94\pms{0.46}} & {6.22\pms{0.30}}
				& {73.78\pms{0.44}} & {6.85\pms{0.36}}
				& {\underline{89.14}\pms{0.23}} & {4.48\pms{0.24}}
				& {~\,75.42\pms{0.55}} & {~~8.65\pms{0.48}} \\
				{SEMA~\cite{SelfExpansionPretrained2025Wang}}
				& {CVPR'25}
				& {78.00\tsb{\qquad~~}} & {5.73\tsb{\qquad~~}}
				& {74.53\tsb{\qquad~~}} & {6.52\tsb{\qquad~~}}
				& {86.98\tsb{\qquad~~}} & {5.35\tsb{\qquad~~}}
				& {~\,80.92\tsb{\qquad~~}} & {~~7.29\tsb{\qquad~~}} \\
				{CL-LoRA~\cite{CLLoRAContinual2025He}}
				& {CVPR'25}
				& {\underline{78.48}\str\tsb{\qquad}} & {5.19\tsb{\qquad~~}}
				& {\underline{76.17}\str\tsb{\qquad}} & {5.95\tsb{\qquad~~}}
				& {87.73\str\tsb{\qquad}} & {5.09\tsb{\qquad~~}}
				& {~\,82.11\str\tsb{\qquad}} & {~~7.17\tsb{\qquad~~}} \\ \midrule
			Mamba-Seq            				& Baseline				 & 63.72\pms{0.55}  	   	  & 28.52\pms{0.49}   		& 56.95\pms{0.62}    	   	& 36.37\pms{0.53}    		& 49.01\pms{0.35}   	& 53.94\pms{0.39}   		& 39.36\pms{0.58}   	& 64.74\pms{0.63}             \\
			Mamba-CL                			& This work              & \textbf{81.67}\pms{0.37}    & 4.07\pms{0.33}      & \textbf{78.60}\pms{0.41}    & 4.88\pms{0.39}      & \textbf{89.60}\pms{0.27}    & 2.57\pms{0.36}      & \textbf{85.03}\pms{0.48}    & ~~3.52\pms{0.54}      \\ \bottomrule
		\end{tabular}
	}
	\label{tab:compare}
\end{table*}

\begin{table*}[t]
	\centering
	\setlength{\tabcolsep}{3pt}
	\caption{Results for long-sequence continual learning under the settings of 50 tasks and 100 tasks across five benchmarks.}
		\begin{tabular}{@{}llcccccccccc@{}}
			\toprule
			\multirow{2}{*}{Method} & \multirow{2}{*}{Venue} & \multicolumn{2}{c}{50S-ImageNet-R} & \multicolumn{2}{c}{50S-CIFAR-100} & \multicolumn{2}{c}{50S-DomainNet} & \multicolumn{2}{c}{100S-ImageNet-R} & \multicolumn{2}{c}{100S-DomainNet} \\ \cmidrule(l){3-12} 
			&           & Acc.          	 & Forgetting    & Acc.          	 & Forgetting         & Acc.          & Forgetting        & Acc.           & Forgetting         & Acc.          & Forgetting         \\ \midrule
			L2P   								& CVPR’22 	& 48.53              & 12.99         & 76.19             & 12.06        & 59.45             & 11.53         & 38.87              & 15.26          & 50.52              & 17.66         \\
			EvoPrompt							& AAAI’24 	& \underline{68.53}  & 10.03         & \underline{76.60} & 13.86        & 67.68             & 10.41         & \underline{61.84}  & 15.84          & \underline{56.35}  & 21.39         \\
			OVOR-Deep							& ICLR’24 	& 60.08              & 5.86          & 65.69             & 14.28        & 66.27             & 7.43          & 40.49              & 8.12           & 47.65              & 8.91          \\
			InfLoRA		 						& CVPR’24 	& 59.02              & 11.02         & 61.49             & 13.68        & \underline{69.96} & 9.51          & 38.16              & 15.11          & 44.32              & 17.85         \\
			EASE  								& CVPR’24 	& 68.17              & 7.76          & 74.47             & 9.31         & 61.20             & 10.01         & 47.55              & 8.22           & 33.08              & 32.14         \\
			CPrompt		 						& CVPR’24 	& 68.47              & 8.16          & 74.97             & 7.45         & 67.87             & 9.36          & 56.95              & 10.20          & 53.73              & 12.14         \\ \midrule
			Mamba-Seq               			& Baseline 	& 19.77              & 72.81         & 11.13             & 85.63        & 4.05              & 94.89         & 7.33               & 79.93          & 2.57               & 81.98         \\
			Mamba-CL                			& This work & \textbf{70.47}     & 4.91          & \textbf{81.05}    & 6.01         & \textbf{72.33}    & 11.69         & \textbf{63.05}     & 8.76           & \textbf{57.91}     & 14.12         \\ \bottomrule
		\end{tabular}
		\label{tab:long-term}
\end{table*}

\begin{table*}[t]
	\centering
	\setlength{\tabcolsep}{5pt}
	\caption{Ablation study of the components proposed in our approach. The first four variables indicate whether the four orthogonal projections in Eq.\eqref{eq: final_projection} for SSM are used, and \( \vH_{out} \) indicates the orthogonal projection applied to the linear layer after the SSM.}
	\begin{tabular}{@{}ccccccccccccc@{}}
		\toprule
		\multirow{2}{*}{\(\vH_{1,\delta}\)} & \multirow{2}{*}{\(\vH_{1,C}\)} & \multirow{2}{*}{\(\vH_{2}\)} & \multirow{2}{*}{\(\vH_{3}\)} & \multirow{2}{*}{\(\vH_{out}\)} & \multicolumn{2}{c}{10S-ImageNet-R} & \multicolumn{2}{c}{20S-ImageNet-R} & \multicolumn{2}{c}{10S-CIFAR-100} & \multicolumn{2}{c}{10S-DomainNet} \\ \cmidrule(l){6-13} 
		&                      &                      &                          &                            	 & Acc.\uar         & Forgetting\dar  & Acc.\uar         & Forgetting\dar  & Acc.\uar         & Forgetting\dar & Acc.\uar         & Forgetting\dar \\ \midrule
		&                      &                      &                          &                           	 & 63.72            & 28.52           & 56.95            & 36.37           & 49.01            & 53.94          & 39.36            & 64.74          \\ \midrule
		\use					&                      &                      &                          & \use                          & 81.08            & 5.38            & 77.47            & 5.97            & 89.15            & 4.13           & 84.19            & 7.09           \\
		& \use                 &                      &                          & \use                          & 79.92            & 7.30            & 76.45            & 8.53            & 87.90            & 6.94           & 83.02            & 8.91           \\
		&                      & \use                 &                          & \use                          & 79.55            & 7.95            & 75.95            & 9.05            & 87.58            & 7.44           & 82.66            & 10.14          \\
		&                      &                      & \use                     & \use                          & 79.75            & 7.82            & 76.38            & 8.76            & 88.12            & 6.76           & 83.05            & 9.20           \\ \midrule
		\use                    & \use                 & \use                 & \use                     &                            	 & 79.42            & 8.25            & 76.23            & 8.91            & 87.79            & 6.98           & 82.89            & 9.31           \\
		&                      &                      &                          & \use                          & 78.25            & 9.14            & 75.53            & 9.73            & 87.28            & 7.67           & 81.98            & 11.26          \\ \midrule
		\use                    & \use                 & \use                 & \use                     & \use                          & \textbf{81.67}   & \textbf{4.07}   & \textbf{78.60}   & \textbf{4.88}   & \textbf{89.60}   & \textbf{2.57}  & \textbf{85.03}   & \textbf{3.52}  \\ \bottomrule
	\end{tabular}
	\label{tab:ablation}
\end{table*}

\begin{table*}[!ht]
	\centering
	\caption{Results of using three different architectures of backbones pre-trained on the ImageNet-1k dataset.}
	\setlength{\tabcolsep}{1.5pt}
	\label{tab:variants}
	\resizebox{\linewidth}{!}{
	\begin{tabular}{@{}ccccccccccc@{}}
	\toprule
	\multirow{2}{*}{Architecture} & \multicolumn{2}{c}{\multirow{2}{*}{Method}}                                                                                                & \multicolumn{2}{c}{10S-ImageNet-R} & \multicolumn{2}{c}{20S-ImageNet-R} & \multicolumn{2}{c}{10S-CIFAR-100} & \multicolumn{2}{c}{10S-DomainNet} \\
	& \multicolumn{2}{c}{}                                                                                                                       & Acc.               & Forgetting    & Acc.               & Forgetting    & Acc.              & Forgetting    & Acc.              & Forgetting    \\ \midrule
	\multirow{3}{*}{ViT}          & \multicolumn{2}{c}{CODA-Prompt~\cite{codaCVPR23}}                                                                                                            & 67.68              & 4.81          & 64.56              & 5.78          & 75.79             & 4.17          & 63.61             & 6.19          \\
	& \multicolumn{2}{c}{CPrompt~\cite{cpromptCVPR24} }                                                                                                                & 59.27              & 5.98          & 53.07              & 4.49          & 71.97             & 7.91          & 69.34             & 5.78          \\
	& \multicolumn{2}{c}{InfLoRA~\cite{infloraCVPR24} }                                                                                                                & 69.47              & 5.63          & 67.42              & 5.93          & 76.56             & 8.68          & 76.82             & 6.34          \\ \midrule
	\multirow{6}{*}{Mamba}        & \multicolumn{1}{c|}{\multirow{2}{*}{\begin{tabular}[c]{@{}c@{}}MambaVision \\ \cite{hatamizadeh2024mambavision}\end{tabular}}} & Mamba-Seq & 64.38              & 25.02         & 64.17              & 26.55         & 65.72             & 33.89         & 62.01             & 37.94         \\
	& \multicolumn{1}{c|}{}                                                                                                          & Mamba-CL  & \textbf{76.18}     & 7.81          & \textbf{73.58}     & 8.74          & 77.41             & 15.30         & 76.14             & 17.21         \\ \cmidrule(l){2-11} 
	& \multicolumn{1}{c|}{\multirow{2}{*}{\begin{tabular}[c]{@{}c@{}}Vim\\ \cite{2024visionmamba}\end{tabular}}}                     & Mamba-Seq & 65.21              & 23.54         & 64.19              & 24.96         & 67.24             & 29.31         & 61.23             & 38.44         \\
	& \multicolumn{1}{c|}{}                                                                                                          & Mamba-CL  & 75.39              & 8.16          & 73.27              & 8.79          & \textbf{78.43}    & 14.34         & 75.68             & 18.12         \\ \cmidrule(l){2-11} 
	& \multicolumn{1}{c|}{\multirow{2}{*}{\begin{tabular}[c]{@{}c@{}}VMamba\\ \cite{2024vmamba}\end{tabular}}}                       & Mamba-Seq & 64.05              & 25.92         & 63.52              & 26.10         & 65.98             & 30.76         & 62.06             & 37.25         \\
	& \multicolumn{1}{c|}{}                                                                                                          & Mamba-CL  & 75.89              & 7.95          & 72.96              & 9.39          & 76.87             & 15.81         & \textbf{77.11}    & 16.64         \\ \bottomrule
\end{tabular}
	}
\end{table*}

\textbf{Evaluation Metrics:}
Following \cite{l2pCVPR22} and most continual learning methods based on pre-training, we report the final average accuracy and the final average forgetting in our paper.
Formally, the final average accuracy and final average forgetting are defined as:
\begin{equation*}
	\begin{split}
		&\mathrm{Accuracy} = \frac{1}{T} \sum_{i=1}^{T}a_{T, i}, \\
		&\mathrm{Forgetting} = \frac{1}{T-1} \sum_{i=1}^{T-1}\max_{j \in \{1, 2, \cdots, T-1\}}(a_{j,i} - a_{T, i}),
	\end{split}
\end{equation*}
where \( T \) is the number of tasks, \( a_{T, i} \) is the accuracy of the \( T \)-th model on the \( i \)-th task data, and \( a_{j, i} \) is the accuracy of the \( j \)-th model on the \( i \)-th task samples.

Note that the aforementioned final average accuracy is the commonly used metric in pre-training-based works.
However, in addition to the final average accuracy, a few works also report mean average accuracy, which is defined as the mean value of the final average accuracies for all tasks.
For a fair comparison, we consistently report the final average accuracy for all the methods in our paper.
We report the mean values of the final average accuracy and final average forgetting over three runs with different random seeds.
Higher accuracy indicates better model performance, while lower forgetting signifies stronger stability (\textit{i.e.}, the ability to retain old knowledge).
{However, since forgetting only measures the performance decrease from each method's own historical best accuracy, a lower forgetting value does not necessarily indicate better overall continual-learning performance. It should therefore be interpreted jointly with the final average accuracy.}
Accuracy is the primary metric that we should focus on, as it reflects the precision of classification in practice.

\begin{table}[t]
	\centering
	\caption{Comparison with existing methods implemented by the same Mamba backbone.}
	\label{tab:backbone}
		\begin{tabular}{@{}lcc@{}}
			\toprule
			\multirow{2}{*}{Method} & \multicolumn{2}{c}{10-split ImageNet-R}            	\\ \cmidrule(l){2-3} 
			& Acc.\uar                 	& Forgetting\dar          	\\ \midrule
			Mamba-fixed            	& 70.39\pms{0.19}          	& 5.44\pms{0.24}         	\\
					{SeqAdapter-Mamba}
				& 	{67.33\pms{0.69}}
				& 	{18.59\pms{0.73}}
				\\
			EASE-Mamba              & 78.13\pms{0.41} 			& 7.17\pms{0.54} 			\\
			Mamba-CL                & \textbf{81.67}\pms{0.37} 	& \textbf{4.07}\pms{0.33} 	\\ \bottomrule
		\end{tabular}
\end{table}

\subsection{Comparison with Existing Approaches}

We compare the proposed method with existing state-of-the-art methods and our baseline (represented by ``Mamba-Seq'') across four benchmarks, as shown in \tablename~\ref{tab:compare}.
Our method Mamba-CL demonstrates an average improvement of {2.0\%} and a maximum improvement of {3.2\%} in accuracy over other leading methods across these four benchmarks, showcasing the superiority of the proposed approach.
Our baseline Mamba-Seq differs from Mamba-CL solely by not using the orthogonal projection matrices for gradient projection, with all other training settings remaining the same.
The proposed Mamba-CL outperforms the baseline by a large margin, enhancing accuracy by 18\%\(\sim\)45\% and reducing forgetting by 24\%\(\sim\)61\% across all benchmarks.
{Although Mamba-CL does not achieve the lowest forgetting in every setting, its objective is not to minimize forgetting in isolation, but to balance previous-task retention and new-task learning. Therefore, the slightly higher forgetting observed in some comparisons reflects the controlled plasticity retained by Mamba-CL rather than insufficient protection of previous knowledge. Overall, Mamba-CL maintains competitive forgetting while achieving the highest final average accuracy across the evaluated benchmarks.}

\begin{figure*}[t]
	\centering
	\includegraphics[]{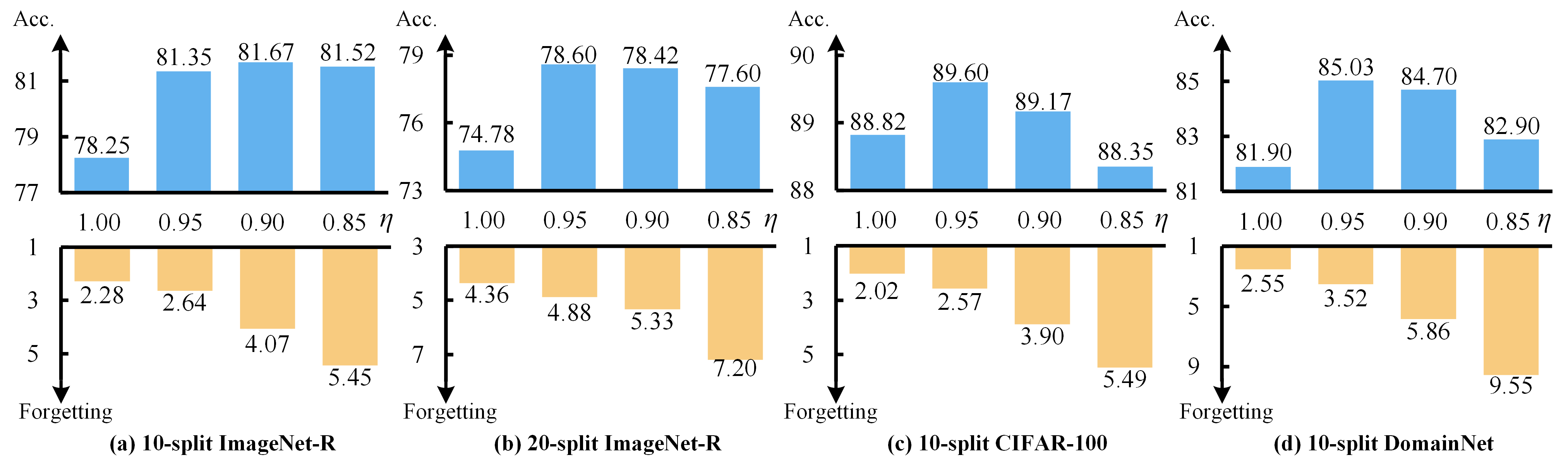}
	\caption{Effects of the orthogonal projection weight \(\eta\) on accuracy and forgetting for the stability-plasticity trade-off.}
	\label{fig:trade-off}
\end{figure*}

\begin{figure}[t]
	\centering
	\includegraphics[]{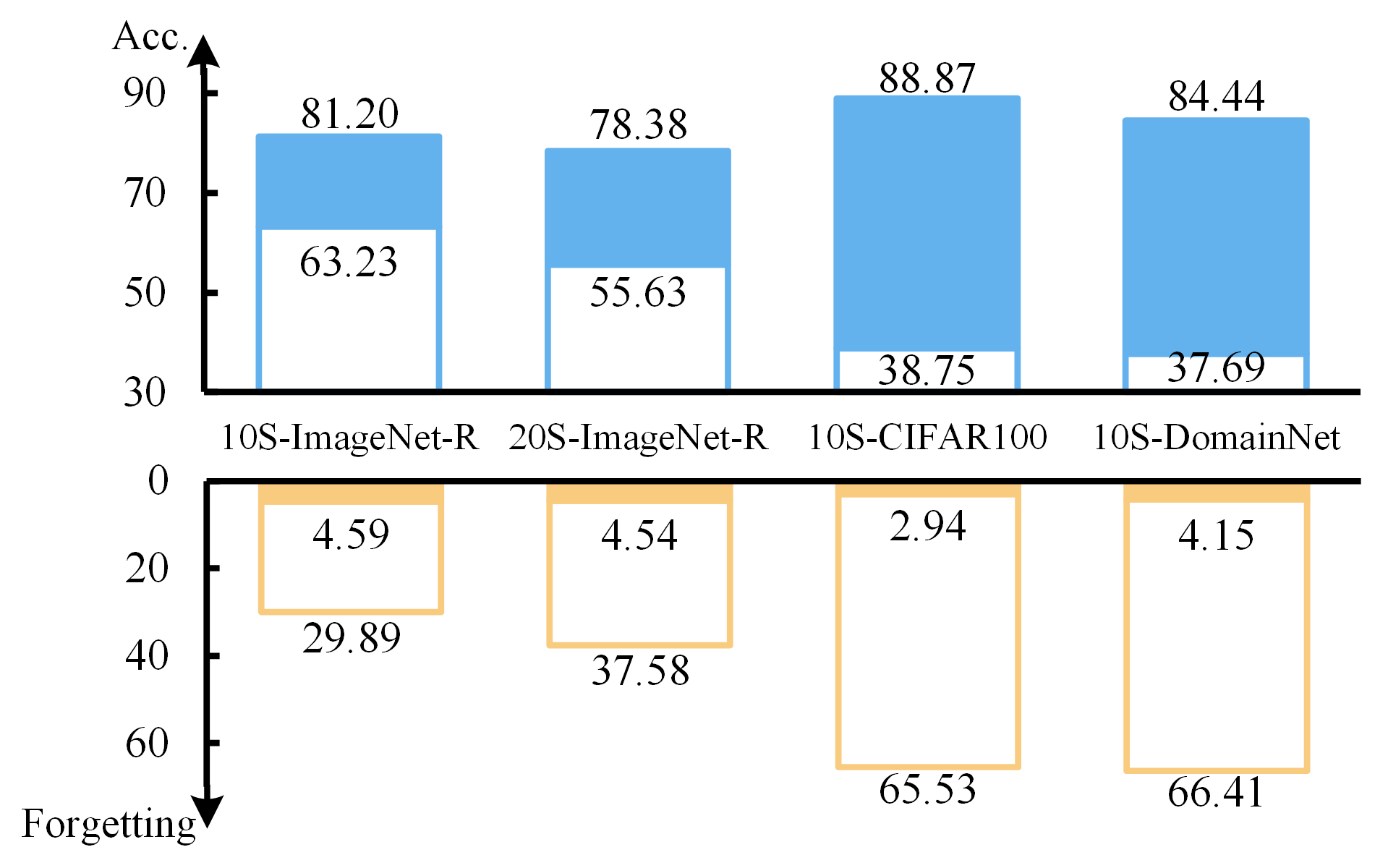}
	\caption{Results of using the pre-training parameters further fine-tuned on ImageNet-1k. The annotated values of the filled bars denote the accuracy or forgetting of Mamba-CL, while those of the blank bars denote the two metrics of the baseline.}
	\label{fig:pretraining}
\end{figure}

To verify that our approach is also suitable for long-sequence continual learning, we conduct experiments across five CIL benchmarks, with task counts reaching 50 and 100, as shown in \tablename~\ref{tab:long-term}.
We additionally reproduce six existing leading methods for comparison.
The results indicate that the baseline Mamba-Seq without anti-forgetting mechanism performs poorly in long-sequence continual learning tasks.
In contrast, Mamba-CL achieves a substantial improvement in accuracy over Mamba-Seq and a significant reduction in forgetting.
Compared with other outstanding methods, Mamba-CL outperforms the second-best approach by an average of 2.3\%, with a maximum improvement of 4.5\%.
This demonstrates that the proposed null-space projection method enables Mamba-CL to maintain its advantage in long-sequence continual learning.

To demonstrate that the superiority of our method arises from the proposed orthogonal projection rather than the Mamba backbone model itself, we further compare our method with {three other} approaches that are also based on the same Mamba backbone.
Specifically, we implement {three} representative methods on the 10-split ImageNet-R dataset, denoted as ``Mamba-fixed'', {``SeqAdapter-Mamba''}, and ``EASE-Mamba''.
Note that since the prompt tuning technique is not widely adopted for fine-tuning Mamba, primarily due to the differences between the backbones of ViT and Mamba, we have chosen not to implement prompt-based approaches (such as L2P \cite{l2pCVPR22} and DualPrompt \cite{dualpromptECCV22}) with Mamba backbone.
As shown in \tablename~\ref{tab:backbone},
``Mamba-fixed'' means that we only train the classifiers, with the Mamba backbone kept fixed.
It is a usual strategy that can prevent forgetting since the features can keep unchanged.
However, the backbone cannot learn new knowledge due to frozen parameters of the backbone.
{``SeqAdapter-Mamba'' freezes the pretrained Mamba backbone and inserts one bottleneck adapter into each of the $S=48$ Mamba blocks. The resulting set of adapters is shared across tasks and sequentially updated without rehearsal, orthogonal projection, or task-specific expansion. The adapter bottleneck dimension is set to $R=16$, while the remaining experimental protocol follows that of Mamba-CL.}
``EASE-Mamba'' is an adapter-based CL approach. It means that we replace the ViT backbone of EASE \cite{easeCVPR24} with the Mamba backbone.
{Although adapter-based methods are lightweight in terms of trainable parameters, adapters do not guarantee effective continual learning.}
{Compared with Mamba-fixed and EASE-Mamba, Mamba-CL improves accuracy by 11.28\% and 3.54\%, respectively. Compared with SeqAdapter-Mamba, Mamba-CL improves accuracy by 14.34\% and reduces forgetting by 14.52\%. These results demonstrate that explicitly protecting previously learned SSM representations is important for achieving stronger continual-learning performance.}

\begin{table}[t]
	\centering
	\caption{Accuracy of training from scratch using the De-focus Mamba-tiny backbone.}
	\label{tab:scratch}
	\begin{tabular}{@{}lcl@{}}
		\toprule
		\multirow{2}{*}{Method}              & \multicolumn{2}{c}{10-split CIFAR100}            \\ \cmidrule(l){2-3} 
		& Acc.\uar      	& Forgetting\dar \\ \midrule
		\multicolumn{1}{c}{Mamba-Seq (tiny)} & 10.78\pms{1.41}     & 21.49\pms{1.73}      \\
		\multicolumn{1}{c}{Mamba-CL (tiny)}  & 32.54\pms{0.82}     & 12.33\pms{0.93}      \\ \bottomrule
	\end{tabular}
\end{table}

\subsection{Ablation Studies}

\begin{figure*}[t]
	\centering
	\includegraphics[]{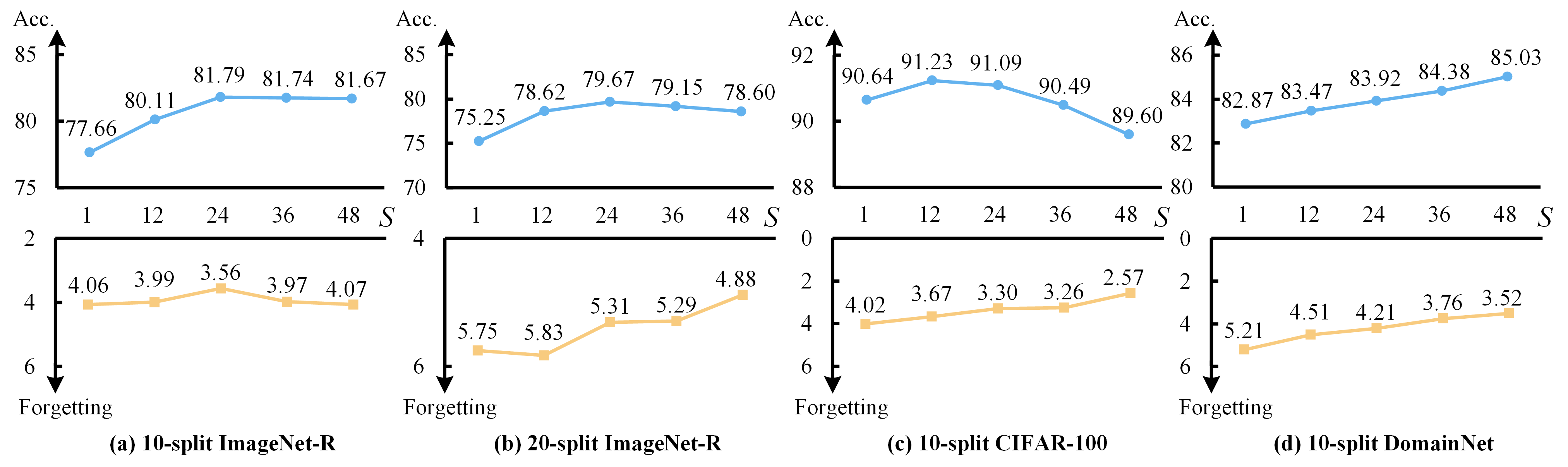}
	\caption{Performance with regard to the number of fine-tuned Mamba blocks (\( S \)). The first \( \{1, 12, 24, 36, 48\} \) Mamba blocks are fine-tuned and the rest blocks are kept frozen.}
	\label{fig:layers}
\end{figure*}

\textbf{Effect of Different Components:}
Our approach comprises three key components, \ie, the three null-space projection matrices \( \vH_{\{1,2,3\}} \).
We use \( \vH_{1,\delta} \) and \( \vH_{1,C} \) to differentiate the \( \vH_{1} \) performed on \( \bm{\delta} \) and \( \vC \), respectively.
As aforementioned, the linear layer after SSM is also trained with a null-space projection matrix, which is denoted as \( \vH_{out} \).
The effect of the introduced projection matrices is shown in \tablename~\ref{tab:ablation}.
We can conclude that: 1) among the four projections for the SSM, the projection of \( \bm{\delta} \) (\ie, \( \vH_{1,\delta} \)) plays the most important role in anti-forgetting.
This is reasonable since \( \bm{\delta} \) is a collective variable that also affects the discretization of \( \vA \) and \( \vB \).
2) The null-space projection applied to the input-invariant parameter $\vA$ (\ie, $\vH_2$) has a relatively minor impact.
3) When the projection matrices are employed jointly, the model achieves the best accuracy with the least forgetting.
This demonstrates that the proposed null-space projections can collectively contribute to Mamba's stability in continual learning.

\begin{figure}[t]
	\centering
	\includegraphics[width=0.95\columnwidth]{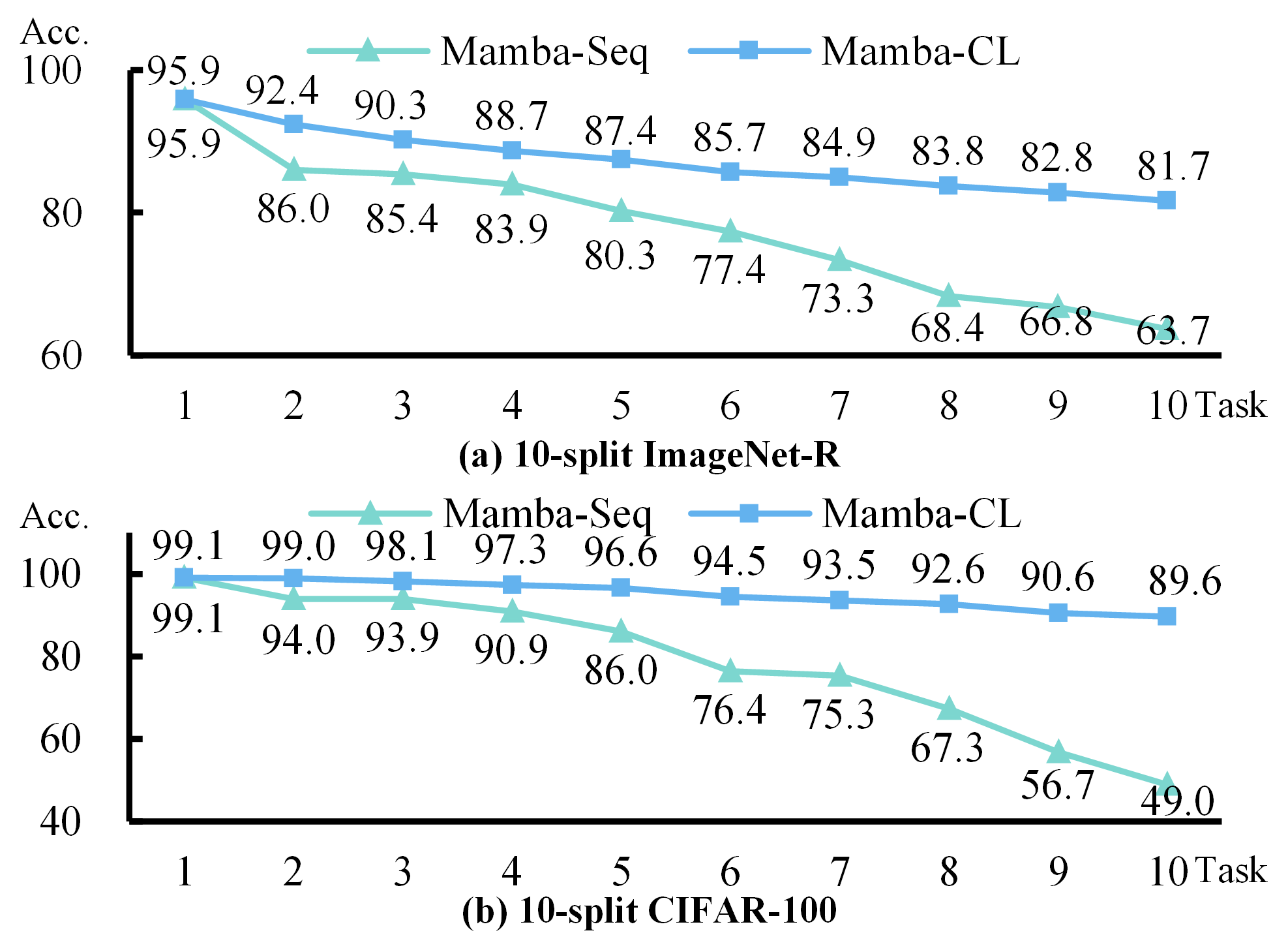}
	\caption{Task-by-task accuracy changing curves of the Mamba-Seq and the Mamba-CL on two benchmarks.}
	\label{fig:acc-curve}
\end{figure}

\textbf{Visual Mamba Variants:}
\label{sec:variants}
To validate that our proposed Mamba-CL is applicable to other visual mamba variants, we conduct experiments on the four benchmarks using the MambaVision~\cite{hatamizadeh2024mambavision}, Vim~\cite{2024visionmamba} and VMamba~\cite{2024vmamba} backbones of their corresponding base versions.
Moreover, we use the official codes to reproduce three ViT architecture-based methods whose backbones are ViT-B/16: CODA-Prompt~\cite{codaCVPR23} (CVPR'23), CPrompt~\cite{cpromptCVPR24} (CVPR'24) and InfLoRA~\cite{infloraCVPR24} (CVPR'24).
All of these Mamba-based and ViT-based models are pre-trained on the ImageNet-1k dataset.
The results are shown in \tablename~\ref{tab:variants}.
It can be seen that our approach improves accuracy by  9\%\(\sim\)14\% across the three Mamba backbones on the benchmarks and reduces forgetting by 15\%\(\sim\)20\%.
The proposed Mamba-CL demonstrates strong generalizability across multiple variants of visual Mambas.
Furthermore, the Mamba-CL method outperforms the three ViT-based methods.
On four benchmarks, the best accuracy achieved under the Mamba architecture surpasses the best accuracy under the ViT architecture by an average of 3.76\%.

\begin{figure}[t]
	\centering
	\includegraphics[]{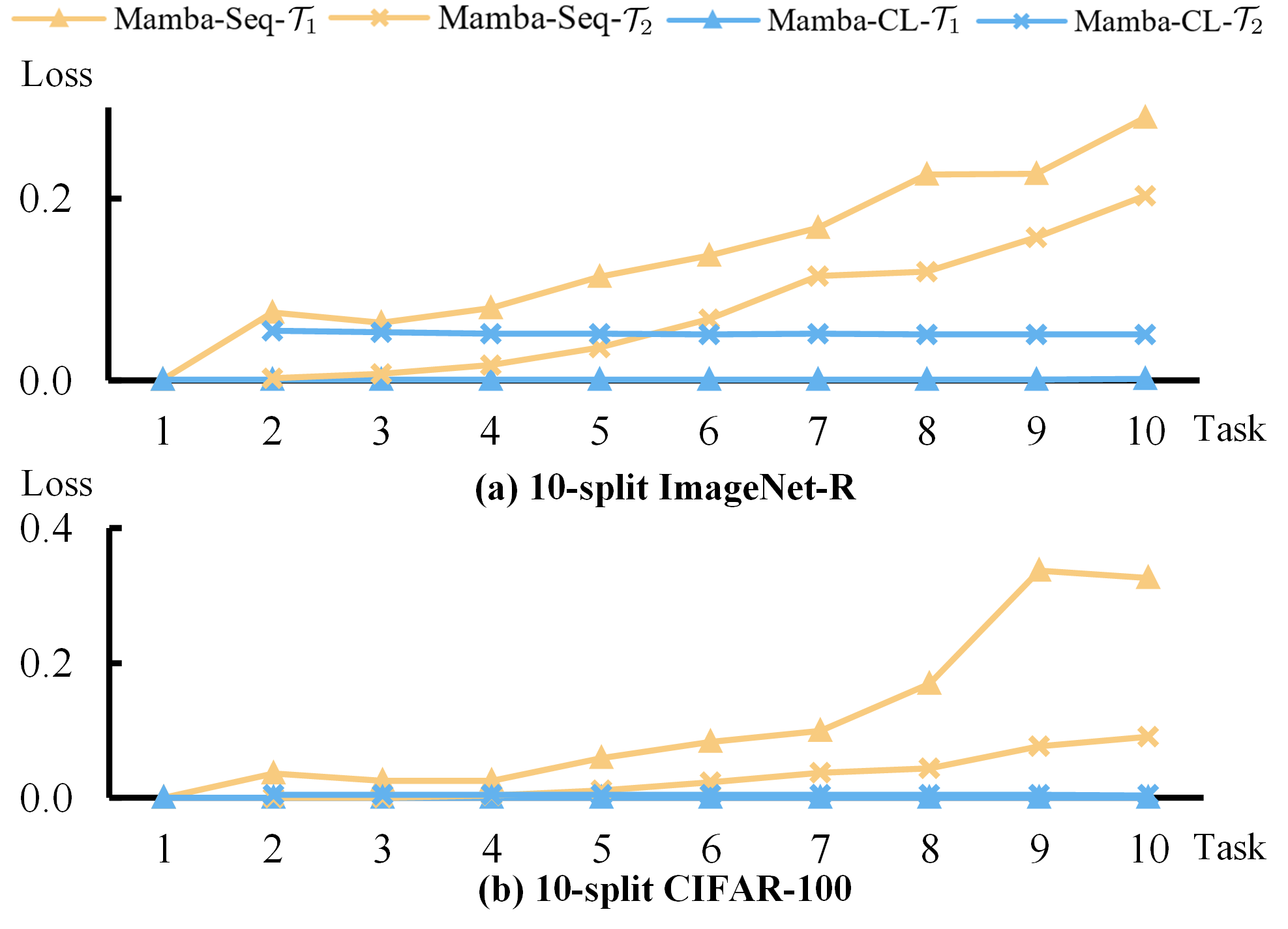}
	\caption{The change of training losses on tasks \(\calT_1\) and \(\calT_2\) of the baseline Mamba-Seq and our Mamba-CL.}
	\label{fig:loss-curve}
\end{figure}

\begin{table*}[t]
	\caption{Comparison of runtime on 10-split ImageNet-R. Variables in ``Complexity'': \( D \): token dimension, \( M \): number of prompts, \( P \): prompt length, \( T \): number of tasks, \( R\): intermediate dimension in the adapter, \( C \): number of classes in a task, \( N \): dimension of each latent state in SSM, \( L \): token length.}
	\centering
	\label{tab:runtime}
		\begin{tabular}{@{}lccccc@{}}
			\toprule
			Method    & Acc.  & GFLOPs & \begin{tabular}[c]{@{}c@{}}Running time \\(minutes)\end{tabular} & \begin{tabular}[c]{@{}c@{}}Inference time \\ (ms)\end{tabular} & Additional Complexity                 \\ \midrule
			CODA-Prompt~\cite{codaCVPR23}	  & 75.46 & 70.4   & 84                                                           & 2.41                                                          & \(\calO(2DM+MPD) \)    \\
			EASE~\cite{easeCVPR24}  		& 76.17 & 35.6   & 95                                                           & 1.87                                                          & \(\calO(T(2DR+R+TCD))\)    \\
			Mamba-Seq & 63.72 & 91.7   & 368                                                          & 5.26                                                          & --                         \\
			Mamba-CL  & 81.67 & 91.7   & 384                                                          & 5.26                                                          & \(\calO(D(2DN+LN+D))\) \\ \bottomrule
		\end{tabular}
\end{table*}

\textbf{Trade-off between Stability and Plasticity:}
As described in Section \ref{sec: null_space_optimizaion}, we introduce a hyper-parameter \( \eta \in [0,1] \) for the null-space projection matrix to balance stability and plasticity.
\figurename~\ref{fig:trade-off} displays the effects of \( \eta \) on accuracy and forgetting.
{Although forgetting is lowest when \( \eta \) is set to 1 (without relaxation), strict projection yields lower final accuracy because it reduces the model's plasticity.}
As \( \eta \) decreases from 1 to 0.85 in steps of 0.05, the accuracy initially increases before decreasing, while the forgetting consistently rises.
This trend indicates a gradual reduction in model stability and an increase in plasticity.
When an optimal balance between stability and plasticity is reached (\ie, \(\eta\)=0.90 for 10-split ImageNet-R and 0.95 for other benchmarks), the model reaches peak accuracy.

{Notably, this stability--plasticity trade-off is not unique to Mamba. A comparable trade-off has also been observed in CNN-based null-space continual learning, where the projection strength must be calibrated to balance previous-task retention and new-task adaptation \cite{TrainingNetworks2025Wang}. However, the present ablation does not establish that SSMs intrinsically require greater plasticity than CNNs or ViTs, because such a conclusion would require a controlled cross-architecture comparison under matched datasets, model capacities, optimization settings, and definitions of projection strength. In Mamba-CL, this trade-off may be particularly visible because strict projection simultaneously constrains the interconnected parameters $\mathbf{A}$, $\mathbf{W}^{\delta}$, $\mathbf{W}^{B}$, and $\mathbf{W}^{C}$ in each SSM block, which jointly govern step-size generation, state transition, input injection, and output readout. Therefore, the results in \figurename~\ref{fig:trade-off} show that Mamba-CL benefits from an appropriately calibrated projection strength, rather than that SSMs inherently require greater plasticity than CNNs or ViTs.}

\textbf{Different Pre-training Parameters:}
We validate that our approach performs well on different pre-training parameters.
As shown in \figurename~\ref{fig:pretraining}, when using the pre-training parameters pre-trained on ImageNet-21k and further fine-tuned on ImageNet-1k, the Mamba-CL can still bring significant accuracy improvements.
Furthermore, we aim to validate that our method can be applied in a training-from-scratch scenario.
We train Mamba-CL and the baseline Mamba-Seq from scratch on 10-split CIFAR-100 with a tiny version of the De-focus Mamba backbone.
In this setting, the parameters in the backbone of Mamba-Seq and Mamba-CL are trained from randomly initialized status.
During training, all the parameters are optimized without any constraints or anti-forgetting techniques for Mamba-Seq.
As to the Mamba-CL, the SSM blocks are still trained with the null-space projections proposed in Eq.\eqref{eq: final_projection}.
The other layers which include the linear layer and convolutional layer are trained using the null-space projections proposed for linear layers \cite{nullspace2021}.
The performance comparison of the Mamba-CL and Mamba-Seq is shown in \tablename~\ref{tab:scratch}.
Mamba-CL still achieves an accuracy improvement of 21.76\%, demonstrating that our approach is applicable to the scenario of training from scratch.

\begin{table*}[t]
	\centering
	\caption{Running time (in minutes) of Mamba-Seq and Mamba-CL on four benchmarks.}
	\begin{tabular}{@{}lccccc@{}}
		\toprule
		Model    	& 10S-ImageNet-R & 20S-ImageNet-R & 10S-CIFAR-100 & 10S-DomainNet & Average   \\ \midrule
		Mamba-Seq  	& 368          & 412          & 715           & 678          & 543      \\
		Mamba-CL  	& 384          & 433          & 755           & 751          & 581      \\
		Increase 	& 16 (4.3\%)   & 21 (5.1\%)   & 40 (5.6\%)    & 73 (10.8\%)  & 38 (7.0\%) \\ \bottomrule
	\end{tabular}
	\label{tab:running-time}
\end{table*}

\begin{table}[t]
	\centering
	\caption{Running time of SVD on 10S-ImageNet-R.}
	\label{tab:svd}
	\begin{tabular}{@{}cc@{}}
		\toprule
		& \begin{tabular}[c]{@{}c@{}} Time (minutes)\end{tabular} \\ \midrule
		Mamba-CL & 384                                                              \\
		SVD      & 8.7 (2.3\%)                                                       \\ \bottomrule
	\end{tabular}
\end{table}

\textbf{Number of Fine-tuned Mamba Blocks:}
There are 48 Mamba blocks in the De-focus Mamba-Large model.
We fine-tune the first \{1, 12, 24, 36, 48\} blocks and freeze the rest blocks.
The accuracy and forgetting are shown in \figurename~\ref{fig:layers}.
The accuracy first increases significantly and then decreases slightly on ImageNet-R and CIFAR-100 datasets.
However, it always increases with the growth of the number of fine-tuned blocks on the 10-split DomainNet.
The reason is that the split DomainNet is a cross-domain CIL benchmark with various domains and classes.
As the capacity of the model increases, the accuracy can improve steadily due to the better adaptation to data.
Since the size of the CIFAR-100 dataset is small and the domains vary little, the model is easy to overfit to it and thus causes slight performance degradation.
Nonetheless, the forgetting can almost keep unchanged or decrease as the number of fine-tuned blocks increases.
It demonstrates that the proposed Mamba-CL can effectively prevent forgetting in each fine-tuned Mamba block.
Given that when all the blocks are fine-tuned, the model can reach the peak accuracy on the DomainNet, and generally achieve satisfactory performance on the ImageNet-R and CIFAR100, we uniformly fine-tune all the Mamba blocks in our experiments.

\subsection{Visualization of Effectiveness}
\figurename~\ref{fig:acc-curve} visually depicts the accuracy curves for the 10-split ImageNet-R and 10-split CIFAR-100.
The performance of the Mamba-CL surpasses the baseline on each task consistently.
Especially, the improvement increases with learning more and more tasks.
To validate the stability of the proposed Mamba-CL on old tasks, we visualize the evolution of training losses on tasks \( \calT_1 \) and \( \calT_2 \) across two benchmarks, as shown in \figurename~\ref{fig:loss-curve}.
Each data point on the curves represents the loss value of the training data in \( \calT_1 \)/\( \calT_2 \) after continuously training Mamba on a new task.
It can be observed that when Mamba is trained with the proposed null-space method, the training losses for these two old tasks remain almost unchanged.
It demonstrates that the representations of previous instances do not change during training new tasks, thus enabling the model to learn new tasks without compromising performance on previously learned tasks, and confirms that the approximation method effectively maintains the model's stability in practice.

\subsection{Resource Consumption}

We first compare our method with CODA-Prompt \cite{codaCVPR23} and EASE \cite{easeCVPR24} in terms of accuracy, FLOPs, running time, inference time, and major complexity.
The comparison on the 10-split ImageNet-R is shown in \tablename~\ref{tab:runtime}.
It can be seen that although our method's inference time is roughly double that of CODA-Prompt, it achieves 6\% higher accuracy.
We will introduce the detailed analysis of memory, complexity and running time in the following parts.

\textbf{Analysis of Memory:}
We analyze the additional memory required by the proposed null-space projections for SSMs as follows.
The introduced uncentered covariance matrices (\textit{i.e.}, \( \overline{\mathbf{X}}_t\overline{\mathbf{X}}_t^{\top} \in \bbR^{D \times D} \), \( \overline{{\delta}}_t\overline{{\delta}}_t^{\top} \in \bbR^{L \times L} \), \( \overline{{\delta}_t\mathbf{X}}_t\overline{{\delta}_t\mathbf{X}}_t^{\top} \in \bbR^{D \times D} \)) and the orthogonal projection matrices (\textit{i.e.}, \( \vH_1 \in \bbR^{D \times D} \), \( \vH_2 \in \bbR^{L \times L} \), \( \vH_3 \in \bbR^{D \times D} \)) need to be stored for calculating and performing orthogonal projections during training.
Therefore, the additional {auxiliary parameters} for a total of $S$ layers is formulated as:
\begin{equation}
	{P_{\mathrm{aux}}} = 2S(2D^2 + L^2).
\end{equation}

In our default setting, \( D=2048 \), \( L=145 \) and $S=48$.
The total additional memory required is approximately 807.32 million floating-point values, which is acceptable since the storage cost is cheap for modern hardware.
Note that this additional memory is constant and does not increase with the number of tasks or classes, providing an advantage in practical applications compared to those approaches requiring to expand networks or store samples for rehearsal.

\textbf{Analysis of Complexity:}
We only consider the additional complexity introduced by the null-space projections in each optimization step.
The complexity of extra operations, such as computing the uncentered covariance matrices and the projection matrices, is omitted, as these are performed only once per task.
For the gradient matrices \( \mathbf{G}^{\delta} \), \(\mathbf{G}^C \), \(\mathbf{G}^A\) and \(\mathbf{G}^B\) which are multiplied by \( \vH_1 \), \( \vH_1 \), \( \vH_2 \) and \( \vH_3 \), respectively, the complexity of matrix multiplication is \( \mathcal{O}(D(2DN+LN+D)) \).
The batch size and epochs are represented as \( n_{batch} \) and \( n_{epoch} \), respectively.
We denote the total number of samples in each task as \( |\mathcal{T}_t| \).
After training each task, the extra complexity introduced by the null-space projections is \(\mathcal{O} ( \frac{ n_{epoch}|\mathcal{T}_t|D(2DN+LN+D)} {n_{batch}} ) \).
For a given dataset where the number of samples \( |\mathcal{T}_t| \) is fixed, and a fixed training setup with constant epochs (\( n_{epoch} \)) and batch size (\( n_{batch} \)), the complexity can be simplified to \(\mathcal{O} ( D(2DN+LN+D) ) \).

\textbf{Running Time:}
We report the average running time over three runs for our Mamba-CL and the baseline Mamba-Seq across the four benchmarks, as shown in \tablename~\ref{tab:running-time}.
Compared to the baseline Mamba-Seq, the running time of Mamba-CL increases by 16\(\sim\)73 minutes (4.3\%\(\sim\)10.8\%) across these benchmarks, with an average increase of 38 minutes (7.0\%).
The additional running time introduced by nulls-space projections is acceptable as it constitutes only a small portion of the overall running time.
Furthermore, we analyze the running time introduced by SVD to demonstrate its minimal cost.
As shown in \tablename~\ref{tab:svd}, SVD operations account for only 2.3\% of the total runtime, with no additional time cost during inference.
The low cost is for the following two reasons: 1) It is performed only once per task to compute the projection matrix.
2) Its computational cost depends on the feature dimension (2048 in our method) and is independent of the dataset size.
Therefore, the SVD introduces negligible computational and time costs for our approach.

\subsection{{Scalability Analysis for Larger Mamba Backbones}}
\label{sec:scalability_larger_mamba}

{We further analyze how the auxiliary storage of Mamba-CL scales when the method is applied to larger Mamba backbones. The complete storage required for the covariance and projection matrices is $M_{\mathrm{aux}}=2S(2D^2+L^2)b$, where $b$ is the number of bytes used to store each value. Since $D^2\gg L^2$ in our setting, the auxiliary storage grows approximately linearly with $S$ and quadratically with $D$. Its asymptotic order, $O(SD^2)$, is the same as that of the dominant dense projection layers in a width- and depth-scaled Mamba backbone.}

{The covariance matrices are used mainly to accumulate feature statistics and update the projection matrices at task boundaries, and can therefore be maintained in CPU memory. Only the projection matrices required for per-step gradient projection need to be stored on the GPU. When stored in FP16, their estimated GPU-resident storage is approximately 0.81\,GB for the default configuration of $S=48$, $D=2048$, and $L=145$, 2.42\,GB for $S=64$, $D=3072$, and $L=145$, and 5.37\,GB for $S=80$, $D=4096$, and $L=145$. Thus, the GPU memory required by the auxiliary states remains explicit and can be determined before training for a given backbone configuration.}

{Moreover, the covariance and projection matrices are updated in place across tasks, so their storage is independent of the number of tasks or classes. They can be discarded after training and introduce no additional inference operations, supporting the practical scalability of Mamba-CL to larger Mamba backbones.}
{Together with the same-backbone comparison in \tablename~\ref{tab:backbone}, this analysis justifies the trade-off introduced by our method: Mamba-CL uses additional training-side optimization state, whose size is fixed for a given backbone and independent of the number of tasks or classes, to obtain stronger continual-learning performance, while avoiding task-specific backbone expansion and additional inference-time projection operations.}

\section{Conclusion}
{In this paper, we introduced Mamba-CL, a structure-aware continual learning framework that reformulates null-space projection for selective state space models. This approach enables continual training of the large-scale Mamba foundation model while aiming to maintain stable outputs from each SSM block for previous-task inputs.} {We derive four conservative sufficient orthogonality conditions for the key time-invariant parameters in the SSM block. Under the adopted decoupled formulation, these conditions provide a tractable approximation to the broader set of consistency-preserving updates and are efficiently implemented through null-space-based gradient projection.} In future work, we aim to explore more flexible constraints on orthogonal projections and extend our method to more variants of Mamba foundation models and applications.

\section*{Data Availability Statement} 

The datasets supporting this research are publicly available through the following repositories:

\begin{itemize}
	\item \textbf{CIFAR-100}: Hosted by the University of Toronto at \url{https://www.cs.toronto.edu/~kriz/cifar.html}
	\item \textbf{ImageNet-R}: Available under MIT License via GitHub repository: \url{https://github.com/hendrycks/imagenet-r}
	\item \textbf{DomainNet}: Provided by Boston University's Multimedia Lab at \url{https://ai.bu.edu/M3SDA/}
\end{itemize}

\noindent These datasets were accessed and utilized in accordance with their respective license agreements and citation requirements.

%
%


\bibliographystyle{spbasic}
\bibliography{reference}    

%
%

\end{sloppypar}
\end{document}